\documentclass[11pt]{article}

\usepackage[margin=1.0in]{geometry} 
\usepackage{microtype}
\usepackage{lmodern}  
\usepackage{setspace}
\usepackage{amsmath,amssymb,amsthm}
\usepackage{algorithm}
\usepackage{algorithmic}
\usepackage{graphicx}
\usepackage{booktabs}
\usepackage{multirow}
\usepackage{subfig}
\usepackage[numbers,sort&compress]{natbib}
\usepackage{array}
\usepackage[table,xcdraw]{xcolor}
\newcolumntype{G}{>{\columncolor{gray!20}}c}

\usepackage[colorlinks=true, allcolors=blue]{hyperref}

\usepackage{authblk}

\def\eqref#1{eq.~(\ref{#1})}
\def\Eqref#1{Eq.~(\ref{#1})}

\def\1{\bm{1}}

\def\rvx{{\mathbf{x}}}

\def\rvz{{\mathbf{z}}}

\def\rmG{{\mathbf{G}}}

\DeclareMathAlphabet{\mathsfit}{\encodingdefault}{\sfdefault}{m}{sl}
\SetMathAlphabet{\mathsfit}{bold}{\encodingdefault}{\sfdefault}{bx}{n}

\def\gA{{\mathcal{A}}}

\def\gD{{\mathcal{D}}}

\def\gL{{\mathcal{L}}}

\def\gO{{\mathcal{O}}}

\newcommand{\E}{\mathbb{E}}

\newcommand{\R}{\mathbb{R}}

\newcommand{\ip}[2]{{\langle #1, \, #2 \rangle}}
\newcommand{\rvX}{\mathbf{X}}
\newcommand{\gLt}{\gL_{\text{total}}}
\newtheorem{theorem}{Theorem}
\newtheorem{assumption}{Assumption}
\newtheorem{definition}{Definition}

\title{\textbf{Controllable Machine Unlearning via Gradient Pivoting}}

\author[1]{Youngsik Hwang}
\author[1,2]{Dong-Young Lim\thanks{Corresponding author: \texttt{dlim@unist.ac.kr}}}

\affil[1]{Artificial Intelligence Graduate School, UNIST, South Korea}
\affil[2]{Department of Industrial Engineering, UNIST, South Korea}

\begin{document}
\maketitle

\begin{abstract}
Machine unlearning (MU) aims to remove the influence of specific data from a trained model. However, approximate unlearning methods, often formulated as a single-objective optimization (SOO) problem, face a critical trade-off between unlearning efficacy and model fidelity. This leads to three primary challenges: the risk of over-forgetting, a lack of fine-grained control over the unlearning process, and the absence of metrics to holistically evaluate the trade-off. To address these issues, we reframe MU as a multi-objective optimization (MOO) problem. We then introduce a novel algorithm, Controllable Unlearning by Pivoting Gradient (CUP), which features a unique pivoting mechanism. Unlike traditional MOO methods that converge to a single solution, CUP's mechanism is designed to controllably navigate the entire Pareto frontier. This navigation is governed by a single intuitive hyperparameter, the `unlearning intensity', which allows for precise selection of a desired trade-off. To evaluate this capability, we adopt the hypervolume indicator, a metric that captures both the quality and diversity of the entire set of solutions an algorithm can generate. Our experimental results demonstrate that CUP produces a superior set of Pareto-optimal solutions, consistently outperforming existing methods across various vision tasks.
\end{abstract}


\section{Introduction}

As the impact and scope of AI models expands across diverse fields, careful data regulation is increasingly required to meet various demands, including user privacy \cite{cao2015towards, ginart2019datadeletion, ghazi2021labeldp}, security \cite{cao2015security, marchant2022hard, di2024hidden}, and fairness \cite{zhang2024fair, oesterling2024fair, chen2024debias}. Machine unlearning (MU), in particular, has emerged as a key approach to ensure compliance with regulations like the ‘right to be forgotten’ in the General Data Protection Regulation (GDPR) \cite{voigt2017gdpr}. Recent advancements in generative models have further underscored the importance of MU in areas such as harmful content mitigation \cite{heng2023selective, gandikota2023erasing, fan2024salun} and copyright protection \cite{dou2024avoiding, yuan2024closer}. 

The most straightforward and exact unlearning approach is to retrain the model from scratch. However, this is computationally intensive and often impractical. Consequently, approximate unlearning methods have gained attention as practical alternatives that form the focus of this work \cite{golatkar2020Fisher, thudi2022unrolling, chen2023boundary, jia2023modelsparsity}.

However, the prevailing approach to approximate unlearning, treating it as a single-objective optimization (SOO) problem by combining efficacy and fidelity into one loss function, creates three primary challenges. \textit{First}, the unlearning process can cause excessive degradation of model fidelity, a phenomenon known as ``over-forgetting'' \cite{thudi2022unrolling}. \textit{Second}, existing algorithms lack a direct mechanism to control the unlearning trade-off, forcing practitioners into an unreliable process of tuning indirect hyperparameters \cite{bae2023gradient, hoang2024gradproj, izzo2021information}. This is a direct symptom of the SOO paradigm, where control is indirect by nature. \textit{Third}, conventional evaluation metrics, which assess a single model's performance, are ill-suited to capture the quality of this trade-off.

In response to these challenges rooted in the SOO paradigm, we propose a fundamental shift to a Multi-Objective Optimization (MOO) perspective. However, we argue that merely adopting an MOO perspective is insufficient, as existing methods are designed to find just a \textit{single} Pareto-optimal solution per run \cite{Yu2020PCGrad, Hwang2025dualcone}. This still fails to provide the practical controllability that users require.

Our core contribution is the \textbf{Pivoting Gradient Principle}, a novel mechanism designed not to find one point, but to controllably \textit{navigate} the entire Pareto frontier. We operationalize this principle in our algorithm, \textbf{CUP} (Controllable Unlearning by Pivoting Gradient), which pivots between two anchors within a conflict-free space. This is governed by a single intuitive ``unlearning intensity'' parameter \(\gamma \in [0,1]\), transforming unlearning from an all-or-nothing operation into a process of surgical precision. To properly evaluate this new capability, we adopt the \textbf{Hypervolume Indicator} as a holistic metric that captures both the quality and diversity of the entire solution set.

In summary, our contributions are:
\begin{itemize}
    \item We reframe machine unlearning as an MOO problem and highlight the limitations of both SOO and traditional MOO approaches for this task.
    \item We propose the Pivoting Gradient Principle and its algorithmic realization, CUP, which enables direct, intuitive, and continuous control over the unlearning process.
    \item We address the shortcomings of traditional evaluation by adopting the Hypervolume Indicator, a metric that holistically captures both solution quality and diversity.
    \item We empirically demonstrate that CUP generates a superior and more diverse set of Pareto-optimal solutions than existing methods across various vision tasks, with comparable computational efficiency.
\end{itemize}

\section{Preliminaries and Problem Formulation}

\subsection{Notations} 

Denote the Euclidean inner product by \(\ip{\cdot}{\cdot}\) and the Euclidean norm by \(\|\cdot\|\). For a vector space \(V\) and its subspace \(W\), the orthogonal complement \(W^\perp\) is defined as:
\[W^\perp := \{v \in V \mid \ip{u}{v} = 0, \quad \forall u \in W \}.\]
For \(v \in V\), we use \(v_{\| W}\) to denote the projection of \(v\) onto a non-zero subspace \(W\). For vectors $\mathbf p,\mathbf q\in \R^{m}$, we denote \(\mathbf{p} \succeq \mathbf{q}\) to indicate that \(\mathbf{p}\) is greater than or equal to \(\mathbf{q}\) in each component, i.e., \(p^{(i)} \geq q^{(i)}\) for all \(i = 1, \dots, m\). Similarly, \(\mathbf{p} \succ \mathbf{q}\) denotes that \(\mathbf{p} \succeq \mathbf{q}\) and there exists at least one index \(j \in \{1, \dots, m\}\) such that \(p^{(j)} > q^{(j)}\). Throughout this paper, let \(\theta \in \Theta \subset \mathbb{R}^d \) denote the parameters of a neural network, where \(\Theta\) represents the parameter space. The gradient with respect to \(\theta\) is denoted by \(\nabla(\cdot)\).


\subsection{Problem Formulation}

Let \( \gD = \{\rvz_i\}^{N}_{i=1} \) be a dataset, where \( \gD_f \subseteq \gD \) denotes the \textit{forgetting dataset} and \( \gD_r = \gD \backslash \gD_f \) denotes the \textit{remaining dataset}. Given a pre-trained model with $\theta_o$, the goal of MU is to adjust the parameters so that the model behaves as if the forgetting dataset $\gD_f$ was never part of the training set.

To formalize this, we define a \textit{forgetting loss} \( \gL_f(\theta) \) 
and a \textit{remaining loss} \( \gL_r(\theta) \) as follows:
\begin{align*}
    \gL_f(\theta) := \E_{\rvX \sim \gD_f}[l_f(\rvX,\theta)],\; \gL_r(\theta) := \E_{\rvX \sim \gD_r}[l_r(\rvX,\theta)],
\end{align*}
where \(l_r: \Theta \times \gD_r \rightarrow \R \), \(l_f: \Theta \times \gD_f \rightarrow \R \) are loss functions for the remaining set and the forgetting set, respectively. Here, the loss functions may vary depending on the specific unlearning tasks and methods. For instance, in classification task, cross-entropy loss \(l_{\text{CE}}\) is typically used \cite{chen2024Null, cha2024instance, jia2023modelsparsity}, while for generative models like diffusion models \cite{ho2020ddpm, rombach2022ldm}, mean squared error (MSE) loss \(l_{\text{MSE}}\) for noise prediction is commonly employed to iteratively refine image samples by denoising noisy inputs \cite{heng2023selective, fan2024salun, li2024I2I}. A common approach in existing approximate unlearning methods is to combine these two competing objectives into a single function, typically a weighted sum~\cite{fan2024salun, jia2023modelsparsity,kim2024layer, li2024I2I, di2024adversarial}:
\begin{align}
    \gL_{\text{total}}(\theta) := w_f \gL_f(\theta) + w_r \gL_r(\theta), \quad \min_{\theta} \gL_{\text{total}}(\theta), \label{eq:single_obj}
\end{align}
where \( w_f, w_r \geq 0\). To solve the single-objective optimization (SOO) in \Eqref{eq:single_obj}, standard gradient descent algorithms such as SGD or ADAM \cite{kingma2014adam} are used, starting from the pre-trained parameters $\theta_o$. This formulation, which collapses the trade-off between efficacy and fidelity into a single objective, represents the \textit{de facto} standard in approximate unlearning. As we will demonstrate, this reliance on the SOO paradigm is the source of fundamental limitations in performance and controllability. 


\subsection{Literature Review} 
Approximate unlearning methods can be broadly categorized by their core strategies.
The most straightforward approaches rely on simple heuristics. For instance, Gradient Ascent (GA) involves inversely training the model on the forgetting dataset by using the negative loss functions \cite{golatkar2020Fisher,thudi2022unrolling}. Random Labeling (RL) neutralizes the forgetting data's influence by assigning it random labels. However, these heuristic methods often suffer from issues of over-forgetting or under-forgetting \cite{bae2023gradient, chen2023boundary, fan2024salun}. To address these limitations, a second category of methods focuses on regularization and strategic parameter updates to enhance their effectiveness. For instance, \cite{jia2023modelsparsity} show that incorporating sparsity can improve the existing unlearning methods, while \cite{fan2024salun} employ saliency-based parameter masking, saliency unlearning (SalUn), which directs updates toward parameters less sensitive to the forgetting dataset. These methods are generally applicable to diverse MU tasks. A third line of research develops methods tailored for specific tasks or model architectures. For image classification, \cite{chen2023boundary} proposes boundary shrink (BS) and boundary expanding (BE) techniques, which effectively shift the decision boundaries of original model. For image generation, most research focuses on diffusion models \cite{schramowski2023safe, zhang2024forget, kumari2023ablating}, with a growing branch of studies dedicated to tasks like concept removal \cite{gandikota2023erasing, kumari2023ablating}. 

\begin{figure*}[!t]
  \centering
  \subfloat[Pre-trained model]{%
    \includegraphics[width=0.24\textwidth]{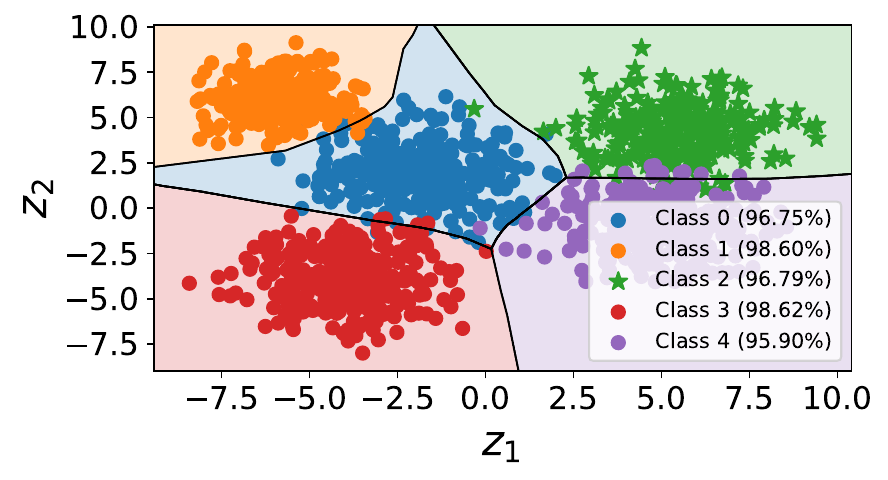}%
    \label{subfig:toy_original}
  }\hfill
  \subfloat[Gradient Ascent]{%
    \includegraphics[width=0.24\textwidth]{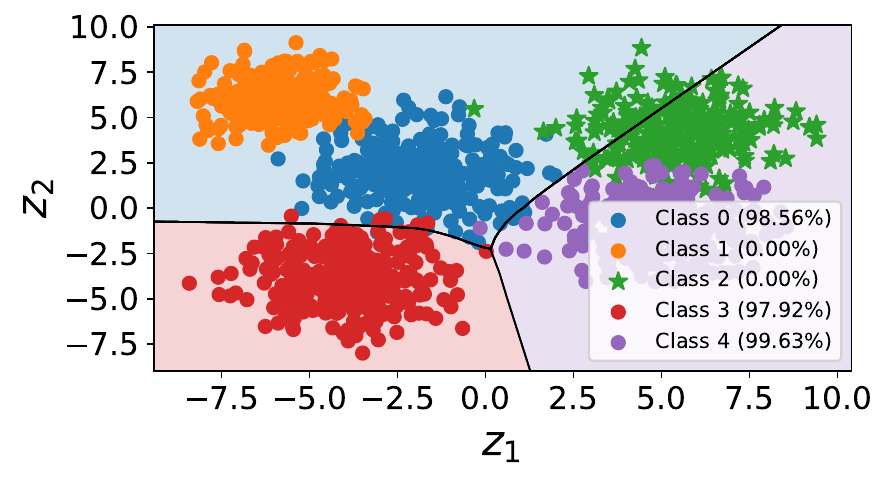}%
    \label{subfig:toy_ga}
  }\hfill
  \subfloat[Weighted Scalarization]{%
    \includegraphics[width=0.24\textwidth]{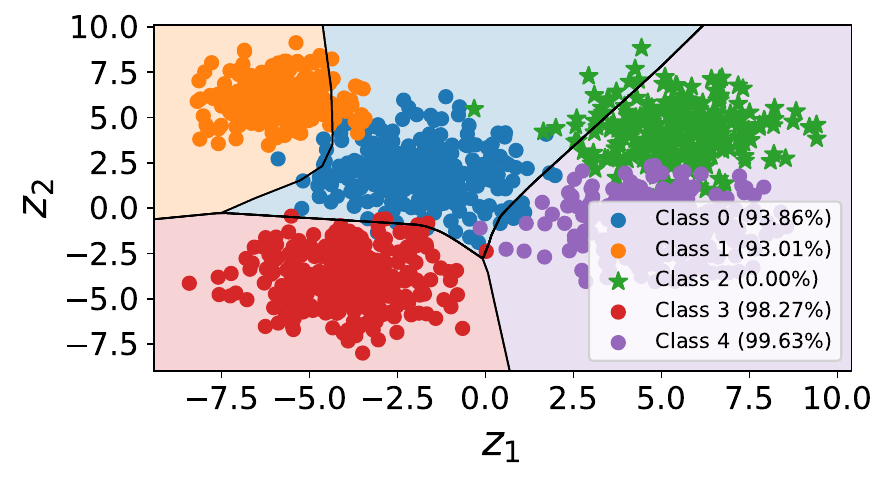}%
    \label{subfig:toy_mtl}
  }\hfill
  \subfloat[CUP]{%
    \includegraphics[width=0.24\textwidth]{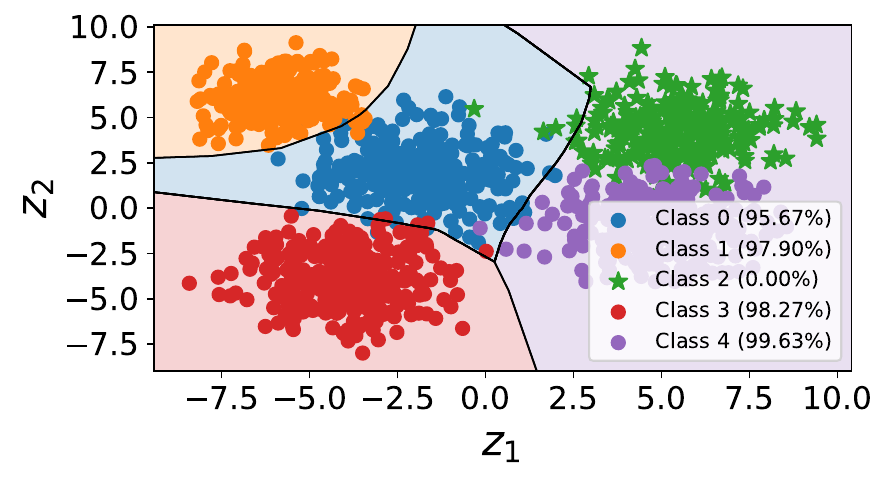}%
    \label{subfig:toy_dcgd}
  }
  \caption{Decision boundaries of a 2-dimensional classification toy example after unlearning Class 2 with different methods: (a) Pre-trained model, (b) Gradient Ascent, (c) Weighted Scalarization, and (d) CUP.}
  \label{fig:toyexample}
\end{figure*}

\section{A Multi-Objective Optimization Perspective on Machine Unlearning}\label{sec:perspective}

This section first discusses the fundamental limitations of prevailing single-objective unlearning methods. We then introduce our core proposal: reframing machine unlearning as a MOO problem. From this new perspective, we define the properties of an ideal unlearning algorithm and propose the hypervolume indicator as a holistic metric to evaluate them.

\subsection{The Pitfalls of Single-Objective Unlearning}\label{subsec:lim}

To illustrate the limitations of existing unlearning algorithms, we consider a toy example of unlearning a classification model initially trained on a mixture of five Gaussian distributions (Figure~\ref{subfig:toy_original}). Our goal is to unlearn Class 2. We refer to Appendix for the detailed experimental setting.

GA adjusts the parameters to intentionally degrade the model’s accuracy on the forgetting dataset by reducing the forgetting loss. This is equivalent to solving the optimization problem in \Eqref{eq:single_obj} with $w_r = 0$. As illustrated in Figure~\ref{subfig:toy_ga}, while GA succeeds in making the model forget the target data, it drastically shifts the decision boundaries for other classes, causing a severe degradation in model fidelity—a phenomenon often called \textit{over-forgetting}.

To mitigate this, several methods implicitly or explicitly minimize a total loss $\gL_{\text{total}}$, which is a weighted sum of the forgetting $\gL_f$ and the remaining losses $\gL_{r}$ as shown in \eqref{eq:single_obj}. We refer to this approach as Weighted Scalarization (WS) \cite{fan2024salun, jia2023modelsparsity, kim2024layer, li2024I2I, di2024adversarial}. While Figure~\ref{subfig:toy_mtl} shows that WS preserve performance on the remaining dataset better than GA, it faces a critical failure when there is a \textit{gradient conflict} between the forgetting loss and the remaining loss. In such cases, minimizing the total loss \(\gLt\) can lead to an increase in one of the individual losses.

This failure can be analyzed by examining the gradients. Suppose the total gradient \(\nabla \gLt(\theta_t)\) conflicts with the remaining loss gradient, i.e., \(\ip{\nabla \gLt(\theta_t)}{\nabla \gL_r(\theta_t)} < 0\). A standard gradient descent step is:
\begin{align}
    \theta_{t+1} = \theta_t - \lambda \nabla \gL_{\text{total}}(\theta_t),
\end{align}
where \(\lambda>0\) is the learning rate. While this update decreases the total loss for a sufficiently small \(\lambda\), the remaining loss \(\gL_r\) inevitably increases:
\begin{align}
&\gL_r (\theta_{t+1}) -\gL_r(\theta_t) \nonumber \\
&=  \ip{\nabla\gL_r(\theta_t)}{\theta_{t+1}-\theta_t} \nonumber + \gO(\|\theta_{t+1}-\theta_t\|^2)\nonumber \\
&= -\lambda \nabla\ip{\gL_r(\theta_t)}{\nabla \gL_{\text{total}}(\theta_t)}+ \gO(\|\theta_{t+1}-\theta_t\|^2)>0. \nonumber
\end{align}
This demonstrates that the existing unlearning process inherently sacrifices model fidelity during gradient conflicts.

Furthermore, all SOO-based methods suffer from a lack of direct control. In practice, unlearning requests often require a specific degree of efficacy. To meet such targets, practitioners must resort to an intensive tuning process of indirect hyperparameters like learning rates, weights (\(w_f, w_r\)), and epochs. This entangled and non-intuitive process makes achieving a desired trade-off challenging and unreliable. These fundamental limitations motivate a paradigm shift in how we approach the unlearning problem.

\subsection{Unlearning as MOO: A Paradigm Shift}\label{subsec:moo_paradigm}

We propose to reframe machine unlearning as a MOO problem. Instead of combining the forgetting loss \(\gL_f\) and remaining loss \(\gL_r\) into a single objective, we treat them as two distinct, competing objectives to be optimized simultaneously. This perspective explicitly manages the trade-off between unlearning efficacy and model fidelity.

While SOO seeks a single minimum, MOO aims to identify a set of \textit{Pareto-optimal} solutions. A solution is Pareto-optimal if one objective cannot be improved without degrading the other. The set of all such solutions forms the \textit{Pareto frontier}. In the context of unlearning, each point on this frontier represents the highest possible model fidelity for a given level of unlearning efficacy. 

Building on this concept, we define the properties of an ideal unlearning algorithm. It must:
\begin{enumerate}
    \item \textbf{Property 1 (Conflict Avoidance):} Simultaneously improve (or at least not worsen) both \(\gL_f\) and \(\gL_r\), even when their gradients conflict.
    \item \textbf{Property 2 (Optimality):} Generate solutions that lie on or very close to the true Pareto frontier.
    \item \textbf{Property 3 (Controllability):} Offer an intuitive mechanism to effectively \textit{navigate} the Pareto frontier and select a desired solution, rather than converging to a single, arbitrary point.
\end{enumerate}

We will introduce our algorithm, CUP, which is explicitly designed to satisfy all three of these properties in Section~\ref{sec:methodology}.

\subsection{Evaluating Controllability: The Hypervolume Indicator}\label{subsec:hypervolume}

Conventional metrics fall short in evaluating algorithms from an MOO perspective. Metrics like Unlearning Accuracy (UA), Remaining Accuracy (RA), and Membership Inference Attack (MIA) assess individual aspects of a single unlearned model \cite{golatkar2020Fisher, chen2023boundary, jia2023modelsparsity, fan2024salun}. This makes holistic comparison ambiguous; if one algorithm yields high RA and another yields high UA, it is unclear which provides a better set of trade-offs. The ``average gap'' to a retrained model, while useful, primarily evaluates optimality (Property 2) but fails to capture the diversity and controllability of solutions (Property 3) \cite{jia2023modelsparsity, fan2024salun}.

\begin{figure}[htb!]
    \centering
    \includegraphics[width=0.8\linewidth]{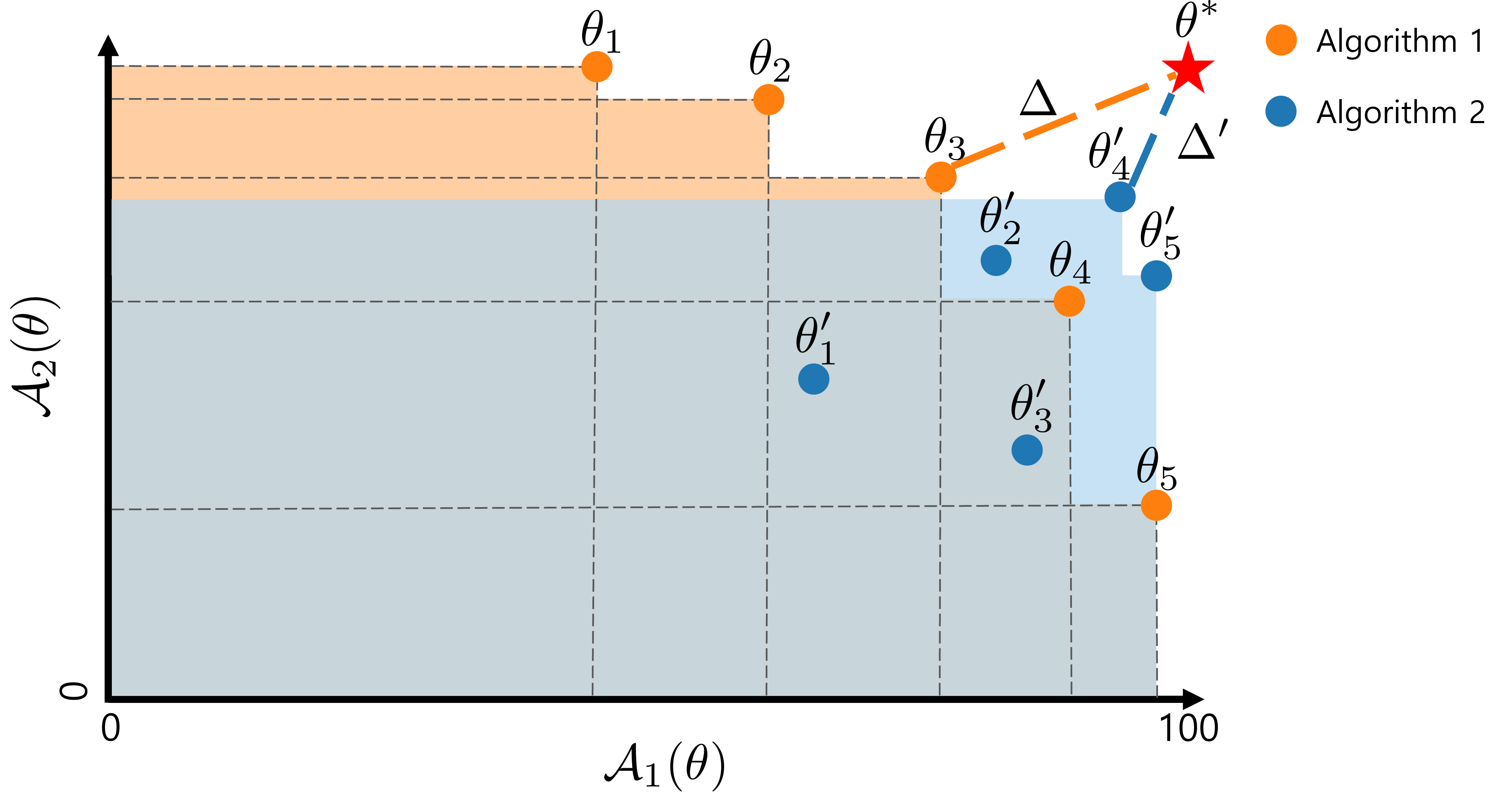}
    \caption{Visualization of hypervolume indicator at the two dimensional performance space. }
    \label{fig:hypervolume}
\end{figure}

To address this, we adopt the \textit{Hypervolume Indicator} from the MOO literature, a rigorous metric that quantifies the volume of the objective space dominated by a set of solutions \cite{zitzler1998multiobjective,guerreiro2021hypervolume}. As illustrated in Figure~\ref{fig:hypervolume}, the hypervolume indicator evaluates both the \textbf{quality} (proximity to the Pareto frontier) and the \textbf{diversity} (spread along the frontier) of the entire solution set an algorithm can generate.

\begin{definition}[Hypervolume Indicator]\label{def:hypervolume}
Let \(\gA(\theta) = (\gA_1(\theta), \dots, \gA_m(\theta)) \in \mathbb{R}^m\) be the evaluation metrics for a parameter \(\theta\), where higher values are better. Given a reference point \(\mathbf{r} \in \mathbb{R}^m\) and a set of parameters \(\mathcal{S} \subset \Theta\), the measure
\[
    \mathcal{H}(\mathcal{S}) = \Lambda \left( \bigcup_{\theta \in \mathcal{S}} \left\{ \mathbf{p} \in \mathbb{R}^m \mid \mathbf{r} \preceq \mathbf{p} \preceq\gA(\theta) \right\} \right)
\]
is the \textit{Hypervolume Indicator} of \(\mathcal{S}\), where \(\Lambda(\cdot)\) is the Lebesgue measure. We set \( \mathbf{r} = \mathbf{0} \).
\end{definition}
Note that each $\gA_i(\cdot)$ can represent any evaluation metric; for example, one might choose $\gA_1$ as RA and $\gA_2$ as MIA. Figure~\ref{fig:hypervolume} provides an intuitive comparison. Algorithm 1, which produces a set of solutions that are both high-quality (closer to the top-right) and diverse (spread out), covers a larger area and thus has a larger hypervolume. In contrast, Algorithm 2 generates solutions that are less diverse, resulting in a smaller hypervolume, even if one of its solutions is locally optimal. A larger hypervolume therefore signifies that an algorithm better satisfies both Optimality (Property 2) and Controllability (Property 3), making it an ideal holistic metric. Unlike traditional metrics, it evaluates the quality and diversity of the entire set of solutions an algorithm can produce. This captures a crucial aspect of practical utility that is overlooked by conventional approaches, which focus on evaluating only a single outcome.

\section{Methodology}\label{sec:methodology}

This section introduces our novel unlearning algorithm, Controllable Unlearning by Pivoting Gradient (CUP). We first derive the core principles for achieving conflict-free updates tailored for machine unlearning. We then present our key contribution, the Pivoting Gradient Principle, which enables precise navigation of the efficacy-fidelity trade-off. Finally, we present the complete CUP algorithm and discuss its novelty.

\subsection{Principle of Conflict-Free Unlearning}\label{subsec:conflict_free}

A fundamental challenge in unlearning is the inherent conflict between the forgetting loss gradient, \(\nabla\gL_f(\theta)\), and the remaining loss gradient, \(\nabla\gL_r(\theta)\). To overcome this, an ideal update vector \(g\) must guarantee that neither loss increases, satisfying the non-negative inner product condition:
\begin{align}\label{eq:conflict_free_condition}
\ip{g}{\nabla\gL_f(\theta_t)} \ge 0 \quad \text{and} \quad \ip{g}{\nabla\gL_r(\theta_t)} \ge 0.
\end{align}
We term the set of all vectors \(g\) that satisfy this the \textbf{Conflict-Free Space}. To derive an operational algorithm, we must identify a tractable subspace, \(\rmG_t\), within this space.

Following the characterization of the conflict-free space in \cite{Hwang2025dualcone}, we construct two critical basis vectors that define the boundaries of \(\rmG_t\). We term these vectors the \textit{Efficacy Anchor} and the \textit{Fidelity Anchor}:
\begin{itemize}
   \item The \textbf{Efficacy Anchor} (\(g_{\text{eff}}\)) is the component of the total gradient orthogonal to \(\nabla\gL_r\), i.e., $\nabla \gL_{\text{total}}(\theta_t)_{\|\nabla \gL_r^\perp}$, representing the conflict-free direction that leans most toward model fidelity.
    \begin{align}
        g_{\text{eff}}(\theta_t) &:= \nabla \gLt(\theta_t) - \frac{\ip{\nabla \gLt(\theta_t)}{\nabla \gL_r(\theta_t)}}{\| \nabla \gL_r(\theta_t) \|^2} \nabla \gL_r(\theta_t). \label{eq:g_r_basis}
    \end{align}
    \item The \textbf{Fidelity Anchor} (\(g_{\text{fid}}\)) is the component of the total gradient orthogonal to \(\nabla\gL_f\), i.e., $\nabla \gL_{\text{total}}(\theta_t)_{\|\nabla \gL_f^\perp}$, representing the conflict-free direction that leans most toward unlearning efficacy.
    \begin{align}
        g_{\text{fid}}(\theta_t) &:= \nabla \gLt(\theta_t) - \frac{\ip{\nabla \gLt(\theta_t)}{\nabla \gL_f(\theta_t)}}{\| \nabla \gL_f(\theta_t) \|^2} \nabla \gL_f(\theta_t). \label{eq:g_f_basis}
    \end{align} 
\end{itemize}
The subspace spanned by the positive combination of these two anchors is defined as \(\rmG_t := \{c_1 g_{\text{eff}}(\theta_t) + c_2 g_{\text{fid}}(\theta_t) \mid c_1, c_2 \ge 0\}\). As shown in \cite{Hwang2025dualcone}, any vector within \(\rmG_t\) satisfies the conflict-free condition in \Eqref{eq:conflict_free_condition}. This provides a rich and computable set of directions for unlearning. However, merely identifying a single direction is insufficient for practical unlearning. This leads to our core proposal: an ideal algorithm must be able to \textit{navigate} the entire spectrum of possibilities between these two anchors.

\subsection{The CUP Algorithm: Navigating the Frontier via Pivoting Gradients}\label{subsec:cup_algorithm}

To enable this navigation of the conflict-free space $\rmG_t$, we introduce the \textbf{Pivoting Gradient Principle}. Our approach defines an update direction that pivots from the fidelity anchor (\(g_{\text{fid}}\)) towards the efficacy anchor (\(g_{\text{eff}}\)). Crucially, this is possible due to the geometric property that all relevant vectors (\(\nabla\gL_f, \nabla\gL_r, \nabla\gLt, g_{\text{fid}}, g_{\text{eff}}\)) lie within the same two-dimensional plane.

We implement this principle by defining the update direction \(g_\gamma\) as a rotation starting from the normalized efficacy anchor, using the pure forgetting gradient \(\nabla\gL_f\) as a reference. This is controlled by the \textit{unlearning intensity} \(\gamma \in [0,1]\):
\begin{align} \label{eq:rotation}
g_\gamma = \cos(\gamma \phi_t)\frac{g_{\text{fid}}}{\|g_{\text{fid}}\|} + \sin(\gamma \phi_t) \frac{\nabla \gL_f (\theta_t)}{\| \nabla \gL_f (\theta_t)\|},
\end{align}
where $\phi_t = \arccos \left( \frac{ \ip{g_{\text{fid}}}{g_{\text{eff}}} }{\|g_{\text{fid}}\| \|g_{\text{eff}}\|} \right)$ is the angle between the two anchors. The coplanar geometry ensures that when \(\gamma=1\), the pivot completes its full arc and \(g_\gamma\) aligns perfectly with the normalized efficacy anchor (\(g_{\text{eff}}\)).

\begin{figure}
    \centering
    \includegraphics[width=0.9\linewidth]{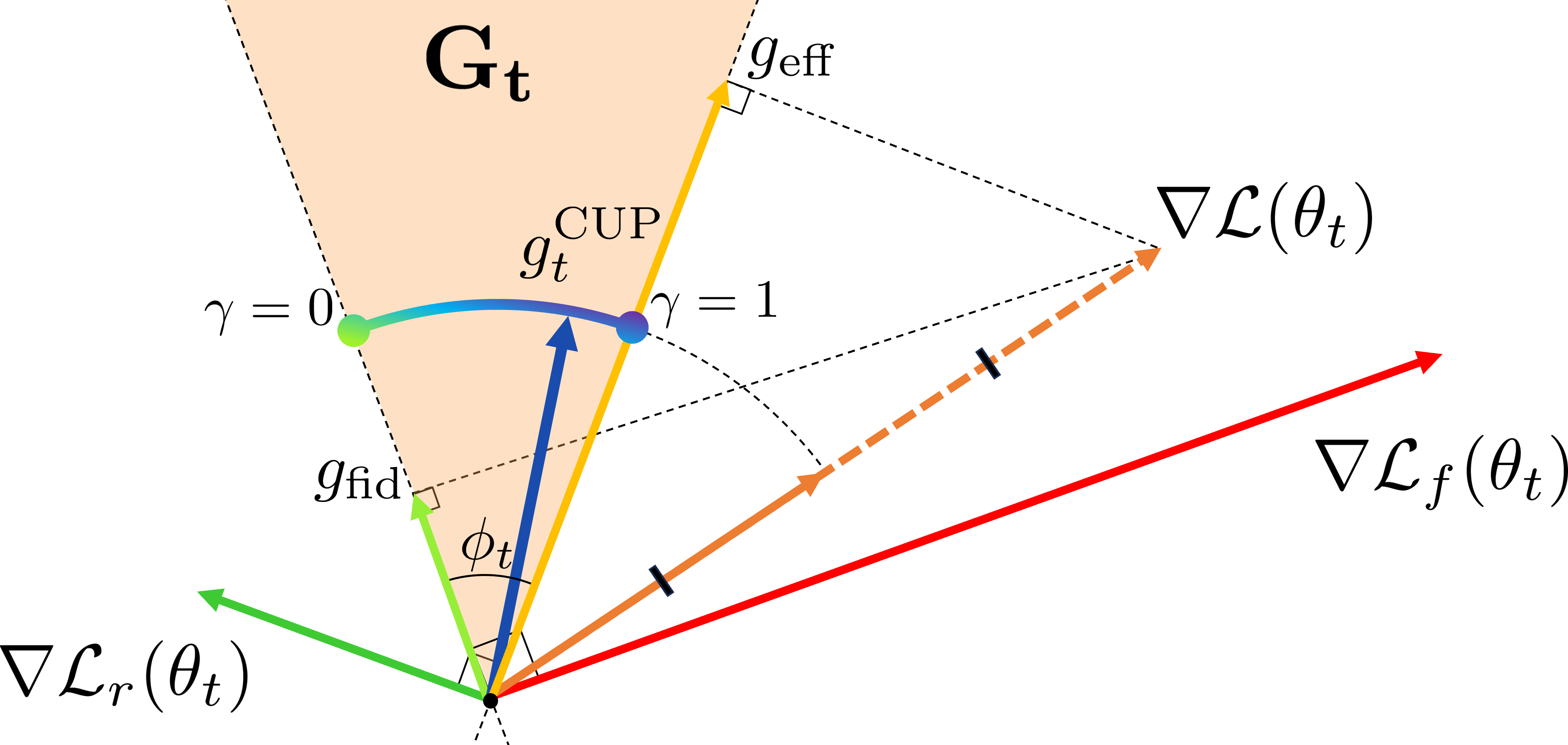}
    \caption{The update vector \(g_t^{\text{CUP}}\) of CUP.}
    \label{fig:CUP}
\end{figure}

This mechanism allows the algorithm to sweep the entire frontier of conflict-free solutions:
\begin{itemize}
    \item When \(\gamma=0\), the direction aligns with the fidelity anchor, prioritizing model fidelity.
    \item When \(\gamma=1\), the direction aligns with the efficacy anchor, prioritizing unlearning efficacy.
    \item For \(0 < \gamma < 1\), the direction smoothly interpolates between these two choices.
\end{itemize}
This establishes \(\gamma\) as an intuitive and predictable control for the unlearning process, allowing practitioners to smoothly navigate the trade-off between efficacy and fidelity. See Figure~\ref{fig:CUP} for the illustration of CUP with varying values of $\gamma$. The final CUP gradient, \(g_t^{\text{CUP}}\), is then scaled by the norm of the total gradient:
\begin{align} \label{eq:cup_update}
g_t^{\text{CUP}} =\| \nabla \gLt(\theta_t) \| g_\gamma.
\end{align}
The complete procedure is detailed in Algorithm~\ref{alg:CUP}.

\begin{algorithm}[htb!]
   \caption{CUP}
   \label{alg:CUP}
\begin{algorithmic}
   \STATE {\bfseries Require:}  learning rate \(\lambda\), max epoch \(T\), initial point \(\theta_o\), unlearning intensity \(\gamma\)
   \STATE {\bfseries Initialize:} $\theta_0 = \theta_o$
   \FOR{\(t=0\) {\bfseries to} \(T-1\)}
   \STATE \(\phi_t = \arccos \left( \frac{ \ip{g_{\text{fid}}(\theta_t)}{g_{\text{eff}}(\theta_t)} }{\|g_{\text{fid}}(\theta_t)\| \|g_{\text{eff}}(\theta_t)\|} \right) \) 
   \STATE \(g_\gamma = \cos(\gamma \phi_t)\frac{g_{\text{fid}}(\theta_t)}{\|g_{\text{fid}}(\theta_t)\|} + \sin(\gamma \phi_t) \frac{\nabla \gL_f (\theta_t)}{\| \nabla \gL_f (\theta_t)\|} \)
   \STATE \( g^{\text{CUP}}_t = \| \nabla \gL(\theta_t) \| g_\gamma \)
   \STATE \(\theta_{t+1} = \theta_{t} - \lambda g^\text{CUP}_t \)
   \ENDFOR
\end{algorithmic}
\end{algorithm}

\subsection{Novelty and Distinction of CUP}\label{subsec:novelty}

CUP's novelty lies not in merely finding a conflict-free direction, but in providing a mechanism to \textit{navigate} the entire frontier of optimal solutions. Existing multi-objective optimizers, including PCGrad \cite{Yu2020PCGrad} and the DCGD \cite{Hwang2025dualcone}, are fundamentally designed to find a single fixed consensus gradient that represents a compromise between objectives. While effective, these methods yield only one point on the Pareto frontier per training run. Consequently, they suffer from the same lack of practical controllability as SOO-based unlearning methods.

CUP overcomes this fundamental limitation through its Pivoting Gradient Principle, operationalized by the unlearning intensity \(\gamma\). This parameter provides unique advantages over conventional hyperparameters, such as learning rates or scalarization weights. Whereas the influence of scalar weights is often complex and indirect, \(\gamma\)'s effect is a direct and geometric rotation, making its impact on the trade-off inherently predictable. This control is also highly intuitive, as \(\gamma\) operates on a fixed \([0, 1]\) scale where \(0\) consistently represents the efficacy-leaning choice and \(1\) represents the fidelity-leaning choice.

This transforms the problem from a brute-force search for a single acceptable model to an elegant exploration of the trade-off space. By providing this level of control, CUP directly satisfies the Controllability (Property 3) of an ideal unlearning algorithm, making it a more practical and user-centric tool.

\section{Theoretical Analysis of CUP}\label{sec:theoretical_analysis}

In this section, we demonstrate that the CUP algorithm is guaranteed to converge to a Pareto-stationary point, where no further improvement can be made to one objective without degrading the other.

\begin{assumption}~\label{ass:lip}
Let the objective functions $\mathcal{L}_f(\theta)$ and $\mathcal{L}_r(\theta)$ be continuously differentiable. The gradient of each objective function, $\nabla\mathcal{L}_f$ and $\nabla\mathcal{L}_r$, is L-Lipschitz continuous. That is, for all $\theta_1, \theta_2 \in \mathbb{R}^d$, there exists a constant $L>0$ such that:
    $$ \|\nabla\mathcal{L}_i(\theta_1) - \nabla\mathcal{L}_i(\theta_2)\| \le L\|\theta_1 - \theta_2\|, \quad i \in \{f, r\} $$
\end{assumption}

\begin{assumption}~\label{ass:compact}
The level set defined by the initial parameter $\theta_0$, denoted as $\mathcal{S} = \{ \theta : \mathcal{L}_f(\theta) \le \mathcal{L}_f(\theta_0) \text{ and } \mathcal{L}_r(\theta) \le \mathcal{L}_r(\theta_0) \}$, is compact.
\end{assumption}

\begin{theorem}[Convergence to Pareto-Stationary Points]
Let Assumption~\ref{ass:lip} and Assumption~\ref{ass:compact} hold. Then every limit point of the sequence of parameters $\{\theta_t\}$ generated by the CUP algorithm is a Pareto-stationary point.
\end{theorem}

\begin{proof}
Let $\{\theta_t\}$ be the sequence of parameters generated by the CUP algorithm. By its construction in Section~\ref{subsec:cup_algorithm}, the update direction $g_t^{\text{CUP}}$ lies within the tractable conflict-free subspace $\rmG_t$. That is, any vector in this subspace satisfies the conflict-free condition from \Eqref{eq:conflict_free_condition}. This guarantees that $\ip{g_t^{\text{CUP}}}{\nabla\mathcal{L}_i(\theta_t)} \ge 0$ for each objective $i \in \{f, r\}$.

We first consider the termination condition. If the point $\theta_t$ is such that for all possible non-zero choices of $g_t^{\text{CUP}}$, we have $\ip{g_t^{\text{CUP}}}{\nabla\mathcal{L}_i(\theta_t)} = 0$ for all $i \in \{f, r\}$, then no descent direction exists within our defined subspace. In this case, $\theta_t$ is a Pareto-stationary point, the algorithm terminates, and the theorem holds.

Therefore, for the remainder of the proof, we focus on the non-trivial case where $\theta_t$ is not Pareto-stationary. This implies that a non-zero update direction $g_t^{\text{CUP}}$ can be found such that $\ip{g_t^{\text{CUP}}}{\nabla\mathcal{L}_j(\theta_t)} > 0$ for at least one objective $j \in \{f, r\}$, while the inner product remains non-negative for the other. From the L-Lipschitz continuity of the gradients (Assumption~\ref{ass:lip}), we have:
$$ \mathcal{L}_j(\theta_t - \lambda g_t^{\text{CUP}}) \le \mathcal{L}_j(\theta_t) - \lambda \ip{\nabla\mathcal{L}_j(\theta_t)}{g_t^{\text{CUP}}} + \frac{L\lambda^2}{2} \|g_t^{\text{CUP}}\|^2 $$
Since the inner product is strictly positive, a sufficiently small step size $\lambda > 0$ must exist that guarantees a strict decrease $\mathcal{L}_j(\theta_{t+1}) < \mathcal{L}_j(\theta_t)$, while ensuring $\mathcal{L}_k(\theta_{t+1}) \le \mathcal{L}_k(\theta_t)$ for $k \neq j$. This establishes that a non-ascending step that strictly improves at least one objective is always possible.

This ensures the sequence of objective vectors $\{F(\theta_t)\}$, where $F(\theta) = [\mathcal{L}_f(\theta), \mathcal{L}_r(\theta)]^T$, is non-increasing in all components and strictly decreasing in at least one. By Assumption~\ref{ass:compact}, the parameter sequence $\{\theta_t\}$ is contained within the compact set $\mathcal{S}$ and thus must have at least one limit point, $\overline{\theta}$.

We now prove by contradiction that $\overline{\theta}$ must be Pareto-stationary. Assume that $\overline{\theta}$ is not Pareto-stationary. By definition, this implies the existence of a descent direction, allowing the CUP algorithm to compute an update $g_t^{\text{CUP}}$ that guarantees a strict improvement $F(\theta_{t+1}) \prec F(\theta_t)$ in a neighborhood of $\overline{\theta}$. This contradicts the fact that the sequence converges to $\overline{\theta}$, as the algorithm would continue to make progress past any such point. Thus, the assumption must be false, and every limit point of the sequence is a Pareto-stationary point.
\end{proof}

\section{Numerical Experiments}\label{sec:main}

\subsection{Experimental Setup}\label{subsec:setup}

\paragraph{Datasets, Models, and Baselines} We evaluate CUP on class-wise forgetting tasks for both image classification and image generation. For classification, we use ResNet-18 \cite{he2016resnet} on the CIFAR-10 \cite{krizhevsky2009cifar10} and SVHN \cite{netzer2011svhn} datasets. For generation, we use a Denoising Diffusion Probabilistic Model (DDPM) \cite{ho2020ddpm} with a U-Net \cite{ronneberger2015unet} architecture on CIFAR-10, following \cite{heng2023selective, fan2024salun}. We compare CUP against a range of baselines, including WS, GA, RL, Influence Unlearning (IU) \cite{izzo2021information}, boundary unlearning (BE and BS) \cite{chen2023boundary}, \(\ell_1\)-sparse \cite{jia2023modelsparsity}, and SalUn \cite{fan2024salun}. For the generation task, we include Selective Amnesia (SA) \cite{heng2023selective} and SalUn. Detailed implementation is available in Appendix. 

\paragraph{Evaluation Metrics}
We evaluate all algorithms based on their ability to satisfy Optimality (Property 2) and Controllability (Property 3). To measure optimality, we compute \( \Delta := \arg \min_{\theta \in \mathcal{S}} \Vert \mathcal{A} (\theta_{\text{retrain}}) - \mathcal{A} (\theta) \Vert_2 \), the minimum distance to the performance of a retrained model. To measure both properties holistically, we use the Hypervolume Indicator \(\mathcal{H}(\mathcal{S})\), as defined in Section~\ref{subsec:hypervolume}. For image classification, the performance vector is \(\mathcal{A}(\theta)=\) (RA, UA, TA, MIA). For image generation, we use FID \cite{heusel2017fid} and a classifier-based accuracy score. To ensure a consistent scale, we rescale FID as \(100 - \min\left(\frac{\text{FID}}{M} \times 100, 100\right)\). We also report run-time efficiency (RTE). The specific hyperparameter search space and evaluation details for each algorithm are provided in Appendix.

\subsection{Controllability of CUP}\label{subsec:controllability_exp}

We now empirically validate its primary advantage: the ability to controllably navigate the efficacy-fidelity frontier (Property 3).

\paragraph{\textbf{Qualitative Demonstration of Control}}
We first demonstrate this control qualitatively. Figure~\ref{fig:ddpm} displays image samples from a DDPM model tasked with unlearning the `automobile' class. As the unlearning intensity \(\gamma\) is increased from 0.1 to 0.5, the characteristic features of an automobile progressively dissolve, eventually becoming unrecognizable. This is achieved while the quality and identity of other classes in the samples remain intact, highlighting CUP's fine-grained control.
\begin{figure*}[!t]
  \centering
  \subfloat[Original]{%
    \includegraphics[width=0.2\textwidth]{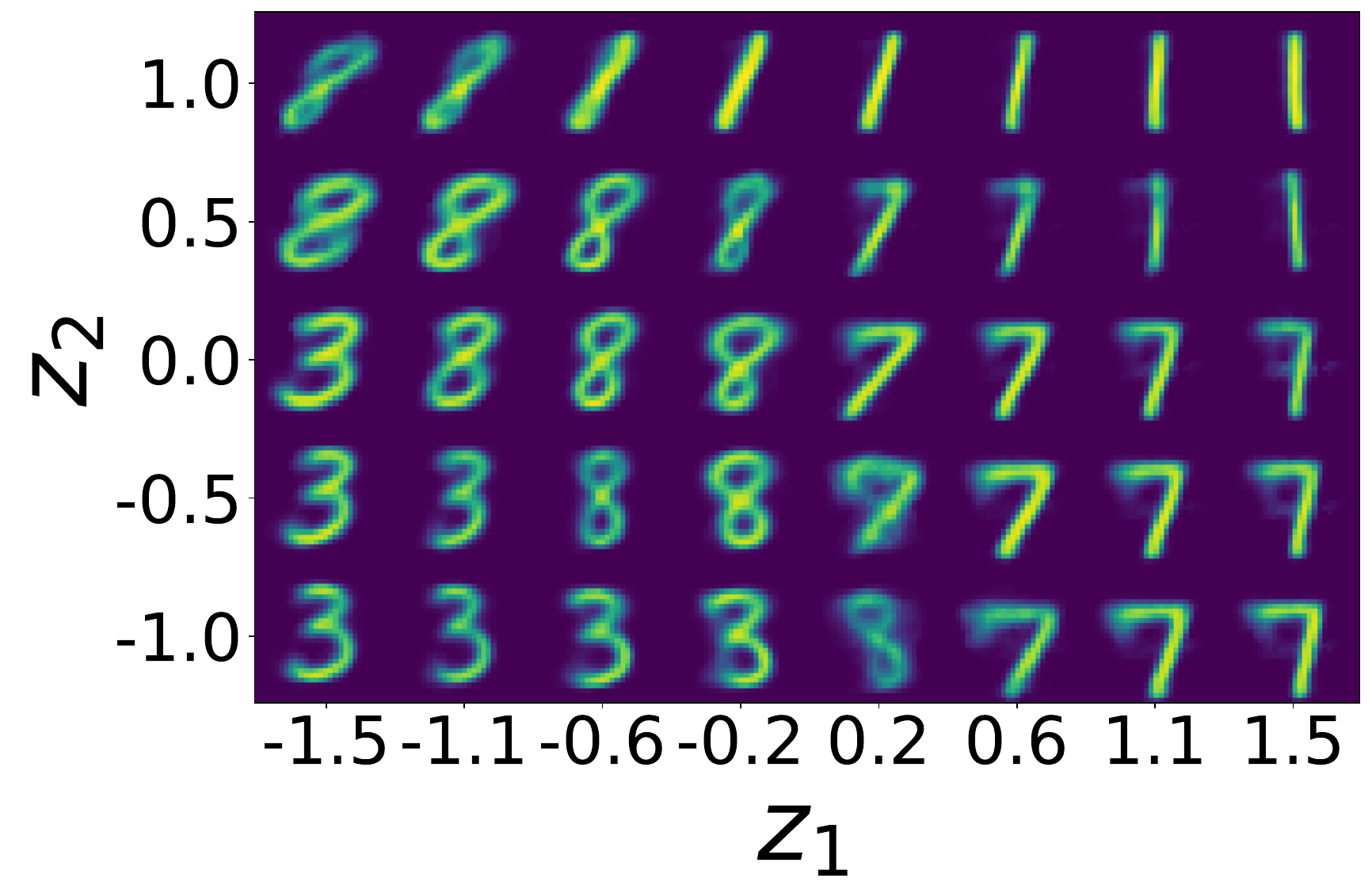}%
  }\hfill
  \subfloat[$\gamma=0.3$]{%
    \includegraphics[width=0.2\textwidth]{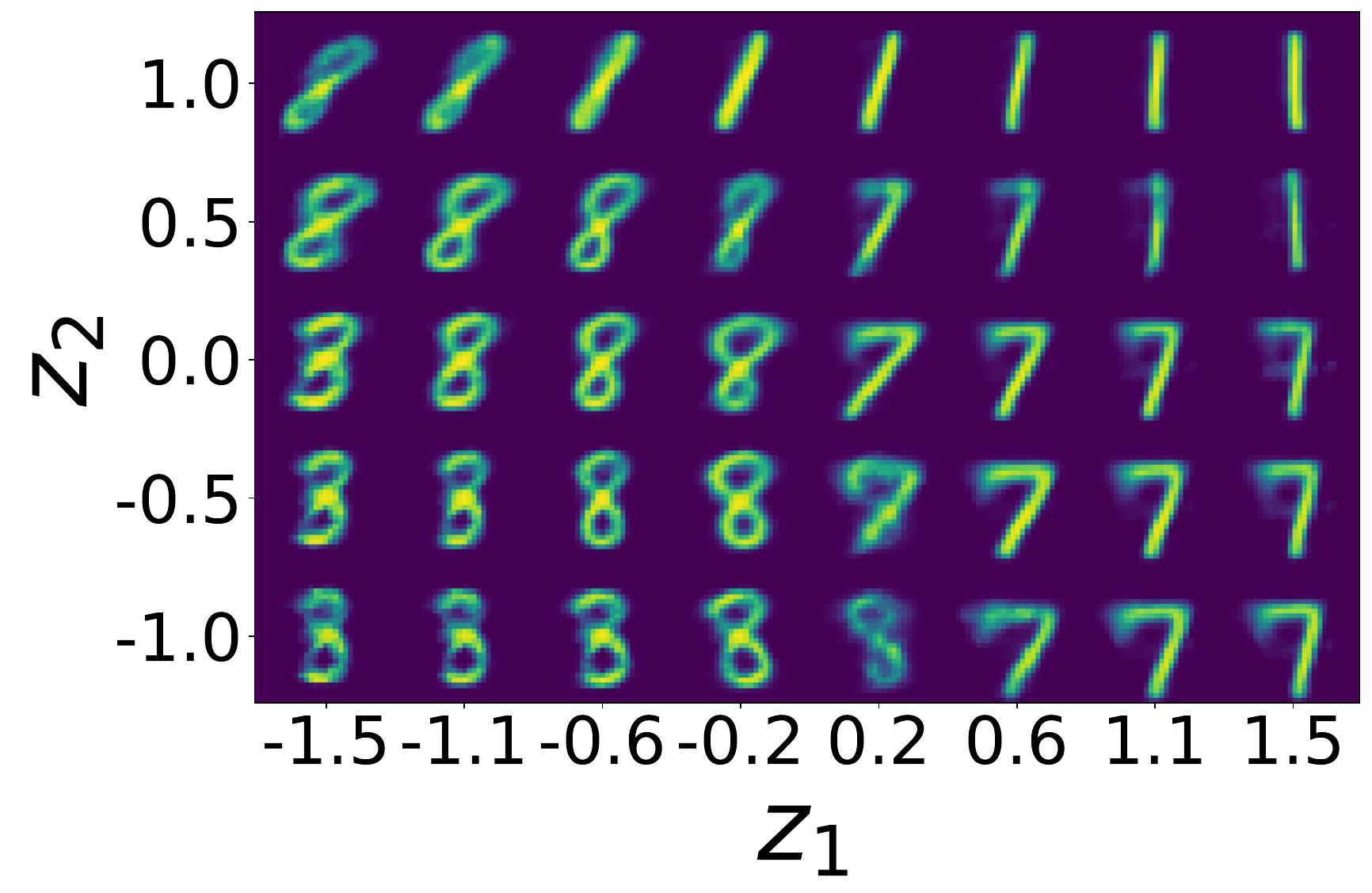}%
  }\hfill
  \subfloat[$\gamma=0.5$]{%
    \includegraphics[width=0.2\textwidth]{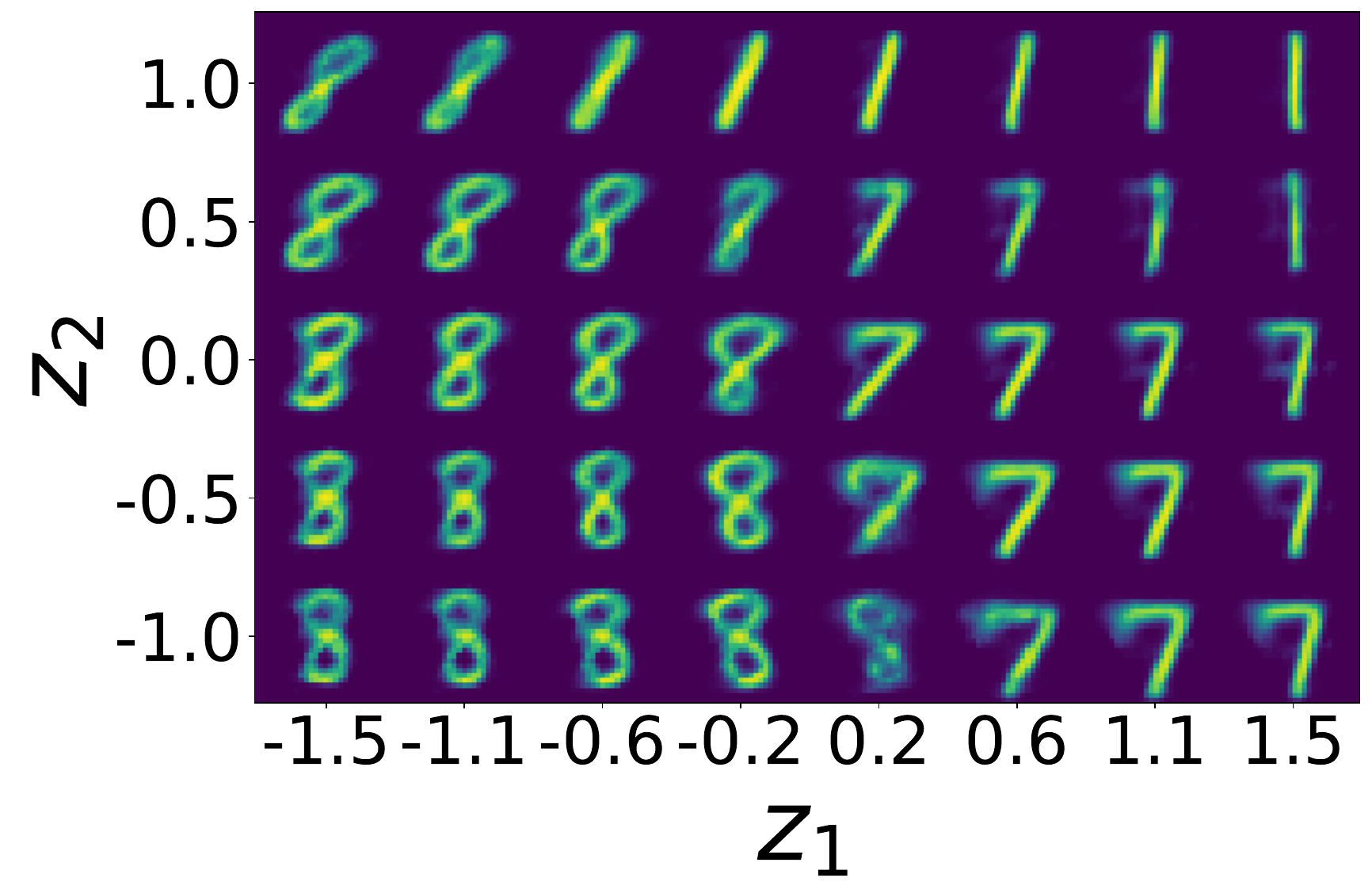}%
  }\hfill
  \subfloat[$\gamma=0.7$]{%
    \includegraphics[width=0.2\textwidth]{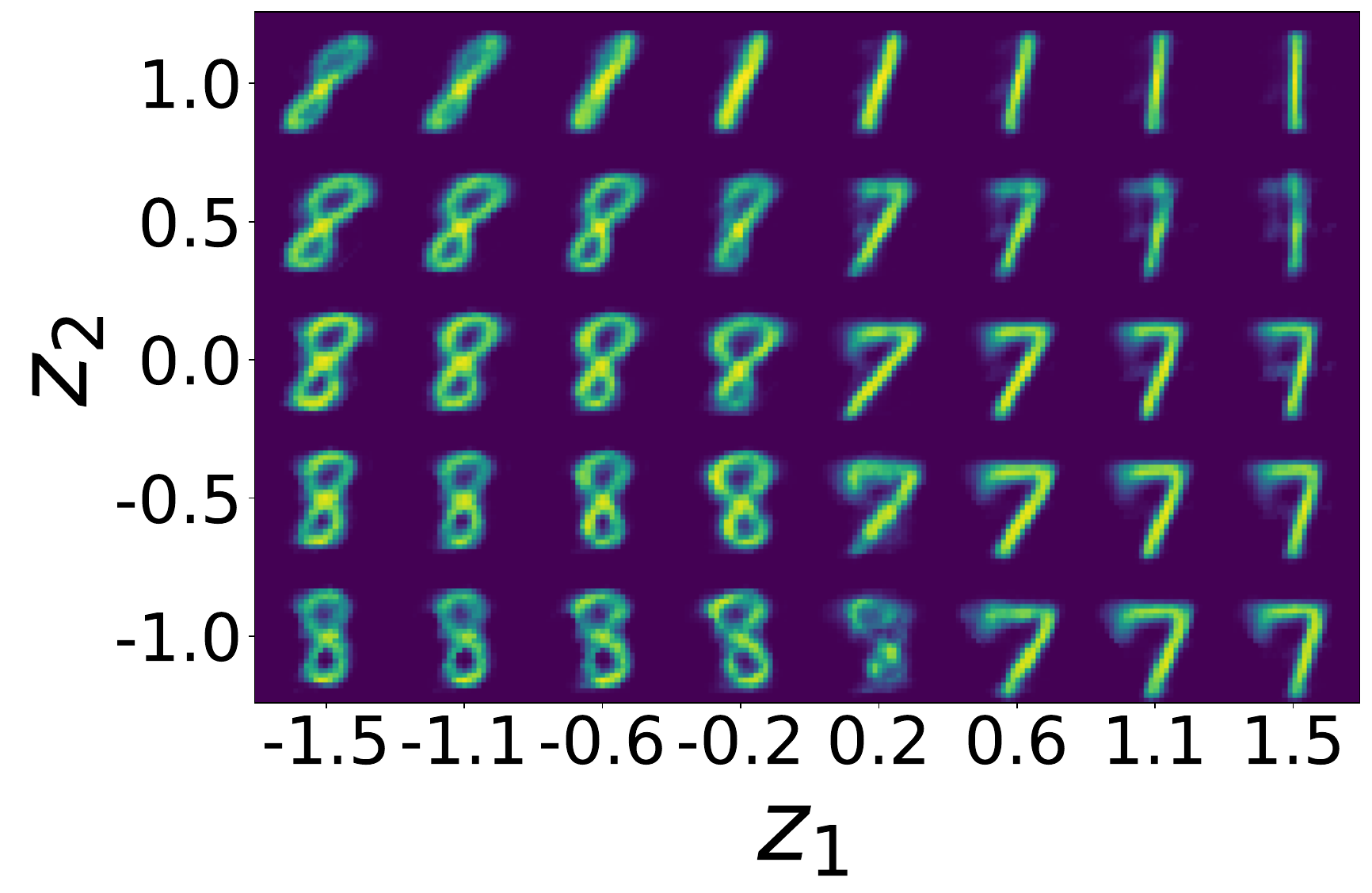}%
  }\hfill
  \subfloat[Retrain]{%
    \includegraphics[width=0.2\textwidth]{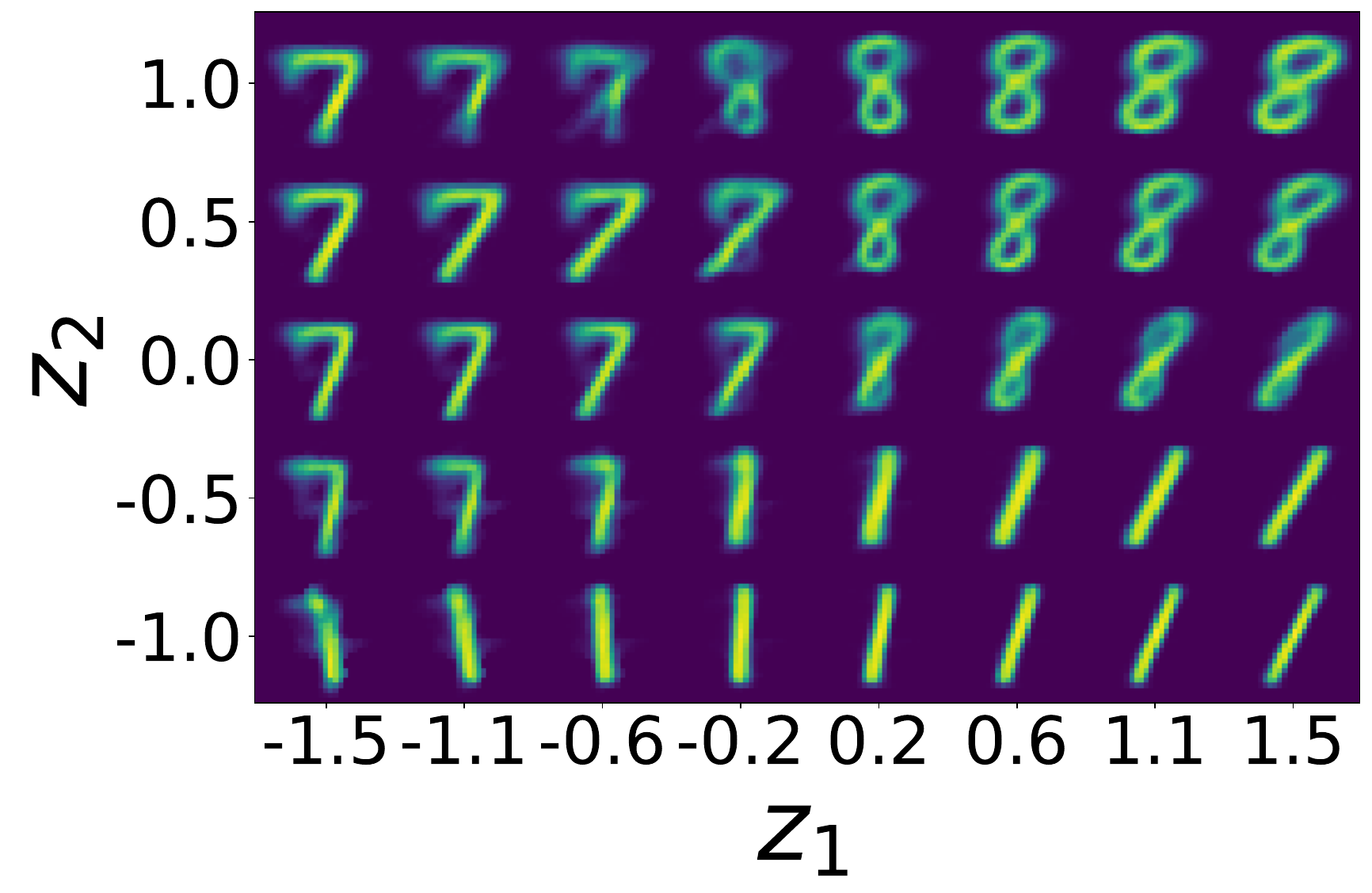}%
  }
  \caption{Visualization of MNIST digit generations from a VAE model with a two-dimensional latent space after unlearning the digit ‘3’. Each point in the $(Z_1, Z_2)$ space corresponds to a generated MNIST digit.}
  \label{fig:mnist}
\end{figure*}

Figure~\ref{fig:mnist} provides another perspective by visualizing the latent space of a VAE trained on MNIST digits, where digit `3' is targeted for removal. As \(\gamma\) increases, the region of the latent space corresponding to digit `3' smoothly transitions to generate images resembling digit `8'. This demonstrates a targeted and gradual removal process that preserves the overall structure of the latent space, in stark contrast to retraining, which leads to a significant and unstructured reconfiguration.

\begin{figure}[!ht]
  \centering
  \includegraphics[width=0.7\textwidth]{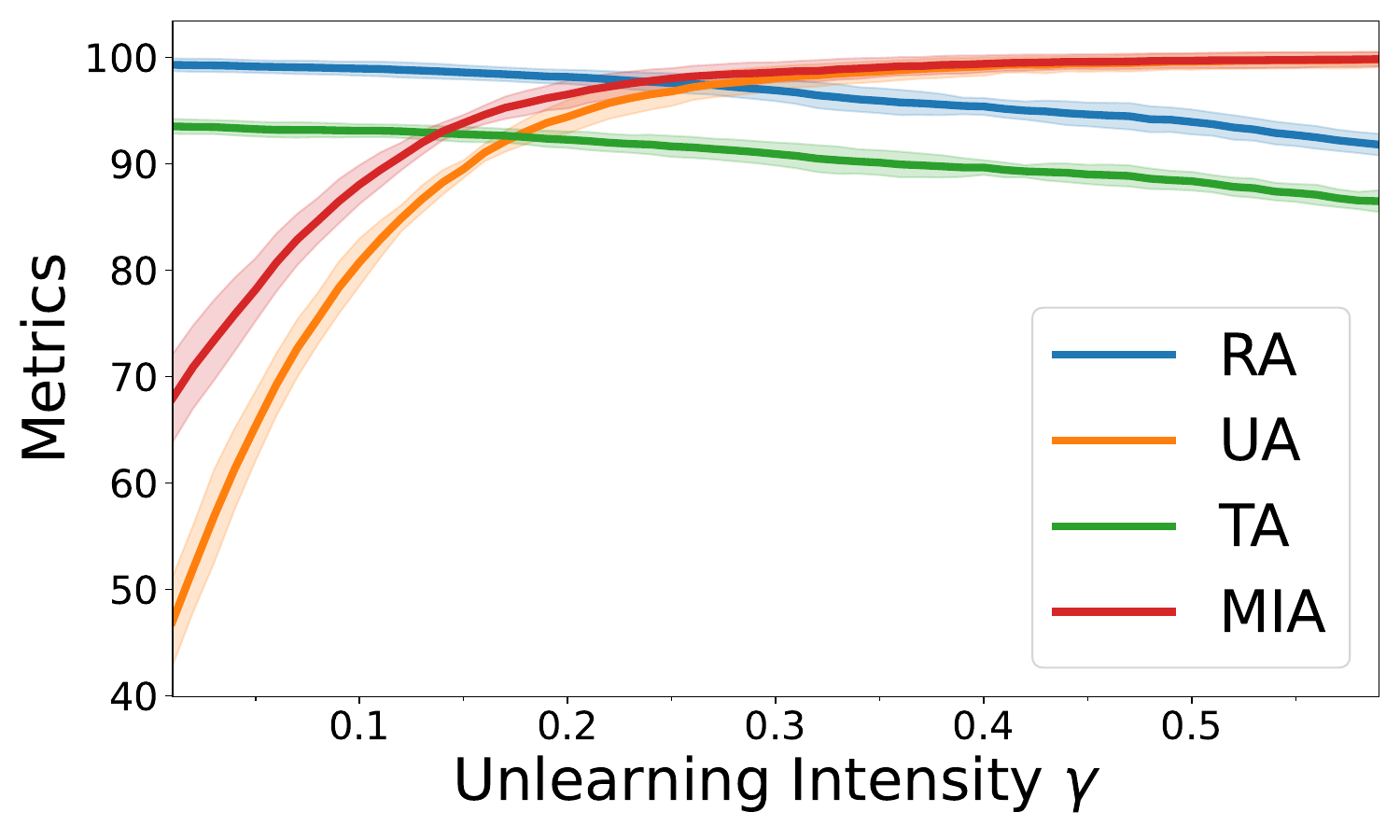}
  \caption{Variation of RA, UA, TA, and MIA with $\gamma$ on CIFAR-10.}
  \label{fig:metrics_w_gamma}
\end{figure}

\paragraph{\textbf{Quantitative Analysis of Control}}
We further verify this controllability quantitatively on the CIFAR-10 classification task. Figure~\ref{fig:metrics_w_gamma} shows the performance metrics as a function of \(\gamma\). As \(\gamma\) increases, the unlearning metrics (UA and MIA) smoothly improve, while the model utility metrics (RA and TA) exhibit a graceful and predictable trade-off. This confirms that \(\gamma\) acts as an effective and intuitive control knob for the degree of unlearning.

\begin{figure}[!ht]
  \centering
  \includegraphics[width=0.7\textwidth]{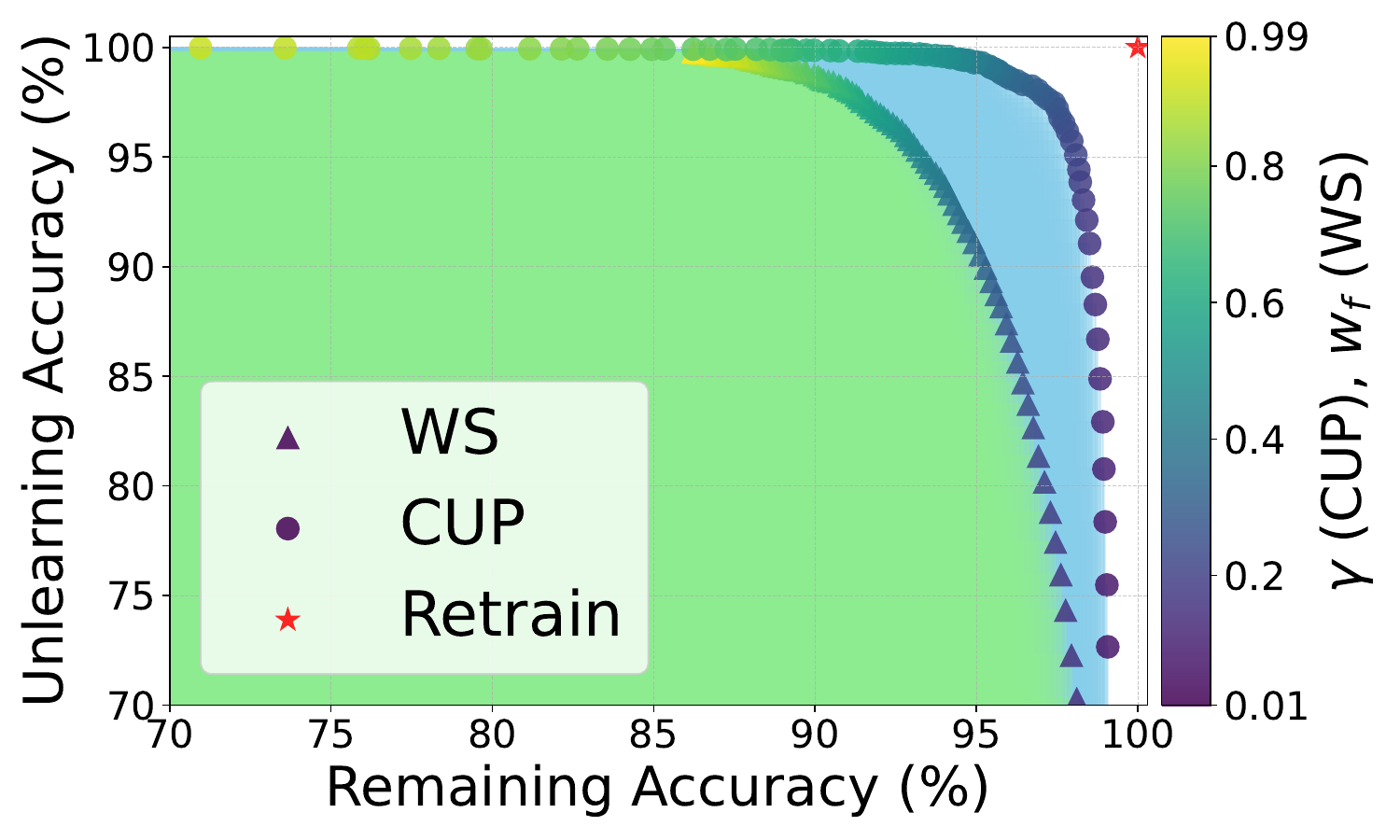}
  \caption{Comparison with weighted scalarization on (RA, UA) space.}
  \label{fig:cup_ws}
\end{figure}

Finally, we compare the set of achievable solutions by CUP against the WS method. Figure~\ref{fig:cup_ws} plots the Pareto fronts of both algorithms in the (RA, UA) space. While WS can generate a range of solutions by varying its weights, the solution set produced by CUP consistently dominates that of WS, achieving higher UA for any given level of RA. This results in a significantly larger hypervolume for CUP and empirically validates its superior ability to generate a better and more diverse set of trade-off solutions, a direct consequence of its ability to navigate a richer solution space beyond the constraints of linear scalarization.

\subsection{Main Results: Image Classification}

\begin{table}[t]
\small
\centering
\caption{Class-wise unlearning performance in image classification task on CIFAR-10 and SVHN. The reported values represent the average performance across 10 classes, with each class’s performance computed over 5 different random seeds. The value within parentheses indicates the class-wise standard deviation.}
\label{tab:classification}
\resizebox{\textwidth}{!}{
\begin{tabular}{l c c c c G G c c c c G G}
\toprule
\multirow{2}{*}{\textbf{Methods}} & \multicolumn{6}{c}{\textbf{CIFAR-10}} & \multicolumn{6}{c}{\textbf{SVHN}} \\ 
\cmidrule(lr){2-7} \cmidrule(lr){8-13}
 & RA ($\uparrow$) & UA ($\uparrow$) & TA ($\uparrow$) & MIA ($\uparrow$) & \cellcolor{white}$\Delta$ ($\downarrow$) & \cellcolor{white}$\mathcal{H}$ ($\uparrow$)
 & RA ($\uparrow$) & UA ($\uparrow$) & TA ($\uparrow$) & MIA ($\uparrow$) & \cellcolor{white}$\Delta$ ($\downarrow$) & \cellcolor{white}$\mathcal{H}$ ($\uparrow$) \\
\midrule
Retrain 
& 100.00 (0.00) & 100.00 (0.00) & 94.88 (0.51) & 100.00 (0.00) & 0.00 (0.00) & 94.88 (0.51)
& 100.00 (0.00) & 100.00 (0.00) & 94.85 (0.63) & 100.00 (0.00) & 0.00 (0.00) & 94.85 (0.63) \\ 
\midrule

GA 
& 94.43 (1.91) & 94.21 (1.33) & 88.31 (1.81) & 96.57 (1.43) & 11.12 (2.40) & 79.90 (3.72)
& 95.86 (3.68) & 95.60 (4.59) & 93.52 (1.54) & 73.20 (42.64) & 29.57 (41.52) & 66.29 (39.67) \\

WS 
& 94.62 (1.07) & 97.38 (1.30) & 88.93 (1.13) & 97.83 (1.09) & 8.83 (1.60) & 87.04 (1.79)
& 97.32 (1.40) & 80.82 (24.73) & 94.11 (1.97) & 85.11 (25.03) & \underline{26.32 (33.69)} & \underline{71.17 (30.59)} \\

RL 
& 91.62 (1.65) & 98.57 (0.90) & 86.35 (1.62) & 98.76 (0.79) & 12.16 (2.12) & 82.63 (2.52)
& 96.93 (1.18) & 81.02 (24.89) & 93.76 (2.13) & 85.20 (25.08) & 26.75 (33.36) & 70.65 (30.56) \\

IU 
& 92.18 (5.24) & 74.68 (0.46) & 86.91 (5.09) & 77.01 (1.13) & 36.52 (2.64) & 55.71 (3.78)
& 95.65 (2.15) & 65.55 (19.47) & 92.26 (2.36) & 83.99 (24.49) & 40.85 (27.75) & 60.70 (24.36) \\

BE 
& 98.97 (0.35) & 79.48 (6.94) & 93.06 (1.00) & 99.21 (0.73) & 20.70 (6.84) & 77.44 (4.50)
& 89.77 (21.12) & 95.66 (4.78) & 87.35 (21.35) & 75.67 (41.72) & 36.66 (45.24) & 62.49 (41.47) \\

BS 
& 98.98 (0.36) & 76.51 (10.40) & 93.12 (0.91) & 98.53 (1.16) & 23.69 (10.34) & 71.97 (10.09)
& 92.79 (10.35) & 74.33 (11.01) & 89.23 (9.35) & 81.17 (27.62) & 37.60 (27.10) & 58.06 (25.17) \\

$\ell_1$-sparse 
& 89.58 (1.06) & 100.00 (0.00) & 86.04 (1.14) & 100.00 (0.00) & 13.67 (1.30) & 80.10 (2.50)
& 82.63 (12.50) & 73.66 (29.19) & 80.84 (10.29) & 68.91 (25.57) & 49.40 (38.49) & 50.28 (33.33) \\

SalUn 
& 96.30 (1.20) & 97.33 (1.82) & 90.60 (1.29) & 98.14 (1.38) & \underline{6.74 (2.13)} & \underline{88.69 (2.03)}
& 96.52 (2.35) & 85.04 (22.70) & 93.70 (2.06) & 83.59 (29.25) & 27.05 (33.59) & 70.67 (30.20) \\

\midrule
CUP 
& 97.79 (0.73) & 98.44 (0.71) & 91.73 (0.96) & 98.94 (0.67) & \textbf{4.34 (1.19)} & \textbf{91.83 (0.97)}
& 97.58 (1.81) & 88.00 (18.98) & 94.66 (1.77) & 86.57 (21.30) & \textbf{19.39 (27.69)} & \textbf{78.47 (25.63)} \\

\bottomrule
\end{tabular}
}
\end{table}

Table \ref{tab:classification} presents the main results for class-wise forgetting on CIFAR-10 and SVHN. For each baseline, we report the performance of the model that achieves the minimum distance to retraining (\(\Delta\)).

The results clearly show that CUP achieves the best performance across the board. On both CIFAR-10 and SVHN, CUP obtains the lowest \(\Delta\) and the highest Hypervolume Indicator (\(\mathcal{H}\)). A low \(\Delta\) value indicates that CUP can produce a single solution that is closer to the gold-standard retrained model than any other approximate method, thereby satisfying Optimality (Property 2). More importantly, the superior \(\mathcal{H}\) score demonstrates that the \textit{entire set} of solutions generated by CUP is of higher quality and diversity, satisfying both Optimality (Property 2) and Controllability (Property 3) more effectively than all baselines.

Furthermore, the results highlight the limitations of evaluating algorithms based on a single metric. For instance, on CIFAR-10, while GA achieves a lower \(\Delta\) than RL, its \(\mathcal{H}\) is smaller. This confirms that merely finding one good solution does not guarantee an algorithm's overall utility, reinforcing the need for a holistic metric like the Hypervolume Indicator.

\begin{figure}[htb!]
    \centering
    \includegraphics[width=0.7\linewidth]{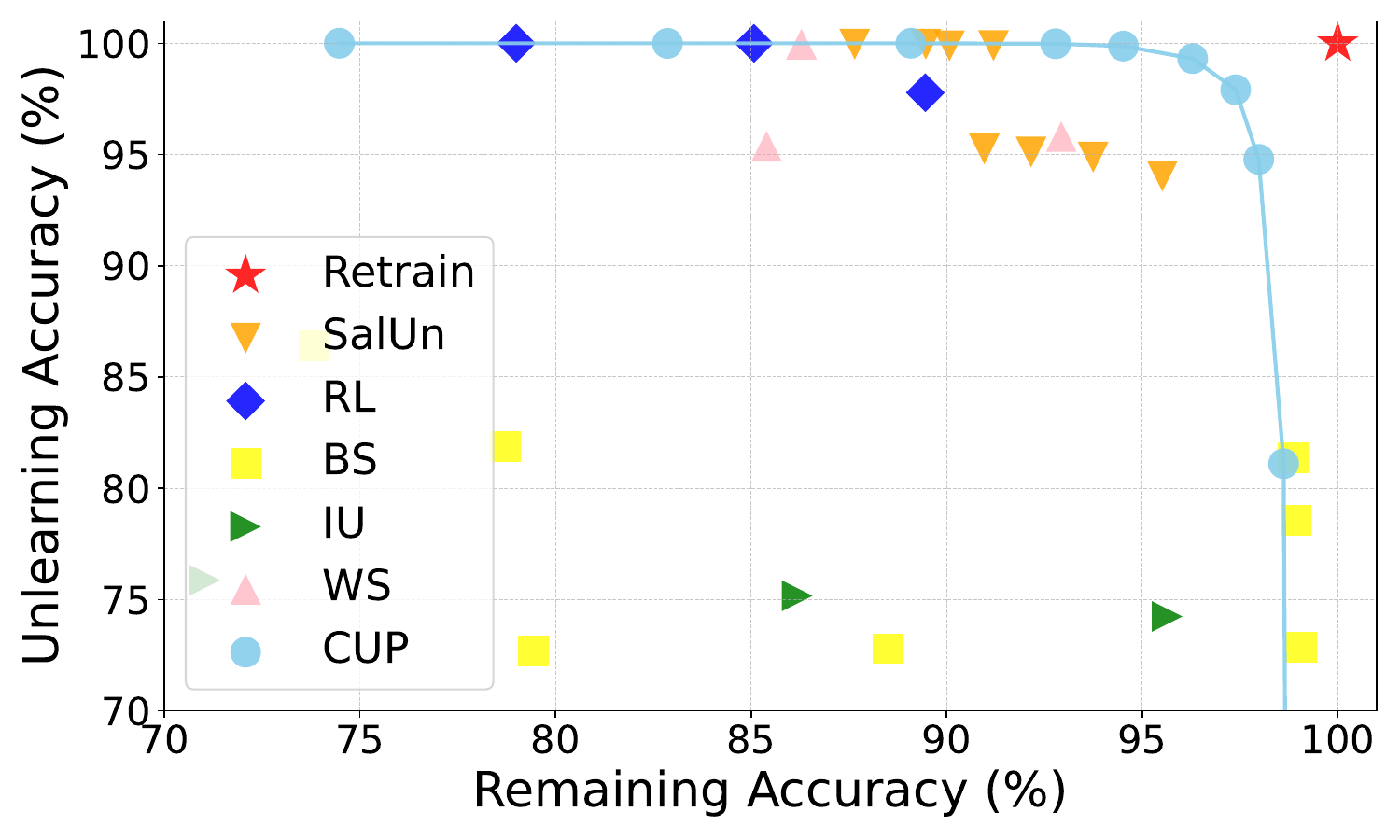}
    \caption{Performance of unlearned models for each unlearning algorithm in image classification on CIFAR-10.}
    \label{fig:pareto front}
\end{figure}

Figure~\ref{fig:pareto front} provides a visual representation of these findings in the (RA, UA) space for CIFAR-10. The solutions generated by varying CUP's \(\gamma\) parameter form a clear and dominant Pareto frontier. In contrast, while a strong baseline like SalUn produces solutions that are close to this frontier, it fails to offer the same level of diversity and control. Other methods generate solutions that are strictly dominated by CUP, lying far from the optimal frontier. These findings empirically support that CUP achieves a superior trade-off between efficacy and fidelity across a wide range of unlearning levels. Finally, as shown in Table~\ref{tab:rte}, this superior performance and controllability do not come at a significant computational cost, with CUP's RTE being comparable to other efficient approximate methods.

\begin{table}[htb]
\small
\centering
\caption{Retraining Time Efficiency (RTE) across methods.}
\label{tab:rte}
\begin{tabular}{lcccccccccc}
\toprule
\textbf{Dataset} & \textbf{Retrain} & \textbf{GA} & \textbf{WS} & \textbf{RL} & \textbf{IU} & \textbf{BE} & \textbf{BS} & $\boldsymbol{\ell_1}$\textbf{-sparse} & \textbf{SalUn} & \textbf{CUP} \\
\midrule
CIFAR-10 & 50.96 & 0.16 & 0.30 & 0.31 & 0.38 & 0.16 & 0.26 & 1.33 & 0.31 & 0.31 \\
SVHN     & 61.21 & 0.13 & 0.25 & 0.25 & 0.46 & 0.13 & 0.46 & 1.60 & 0.25 & 0.25 \\
\bottomrule
\end{tabular}
\end{table}

\subsection{Main Results: Image Generation}\label{subsec:main_results_gen}

We further evaluate CUP on the more challenging task of class-wise unlearning for a generative DDPM model. Table~\ref{tab:gen} reports the performance of CUP against strong baselines, SA and SalUn, on CIFAR-10. The results demonstrate CUP's exceptional efficiency and performance in this domain.

Across multiple classes, CUP achieves \(\Delta\) and \(\mathcal{H}\) scores that are either the best or highly competitive with SalUn, a state-of-the-art method for this task. This indicates that CUP can produce high-quality, diverse sets of unlearned generative models that are comparable to specialized methods.

\begin{table}[ht]
\small\centering
\caption{Performance of class-wise forgetting in image generation on CIFAR-10.}\label{tab:gen}
\begin{tabular}{lcccccc}
\toprule
\multirow{2}{*}{\textbf{Methods}}  & \multicolumn{2}{c}{SA} & \multicolumn{2}{c}{SalUn} & \multicolumn{2}{c}{CUP} \\
\cmidrule(lr){2-3} \cmidrule(lr){4-5} \cmidrule(lr){6-7}
 & \(\Delta\) (\(\downarrow\)) & \(\mathcal{H}\) (\(\uparrow\)) &\(\Delta\) (\(\downarrow\)) & \(\mathcal{H}\) \((\uparrow\)) & \(\Delta\) (\(\downarrow\)) & \(\mathcal{H}\) (\(\uparrow\)) \\ \midrule
Airplane   & 7.10 & 96.39 & 3.52 & 96.49 & \textbf{3.28} & \textbf{96.70}  \\
Automobile & 20.42 & 83.67 & 3.60 & 96.39 & \textbf{3.55} & \textbf{96.52}  \\
Bird       & 9.51 & 94.77 & \textbf{3.58} & \textbf{96.41} & 3.86 & 96.15 \\
Cat        & 14.63 & 88.54 & \textbf{3.62} & \textbf{96.38} & 3.78 & 96.31 \\
Deer       & 5.65 & 94.28 & 3.80 & 96.20 & \textbf{3.53} & \textbf{96.48}  \\ \midrule
RTE        & \multicolumn{2}{c}{89.67} & \multicolumn{2}{c}{13.11} & \multicolumn{2}{c}{6.89} \\
\bottomrule
\end{tabular}
\end{table}

\begin{figure*}[!t]
  \centering
  \subfloat[$\gamma=0.1$]{%
    \includegraphics[width=0.19\textwidth]{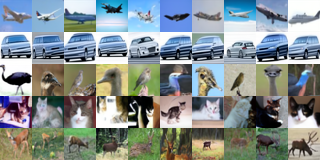}%
  }\hfill
  \subfloat[$\gamma=0.2$]{%
    \includegraphics[width=0.19\textwidth]{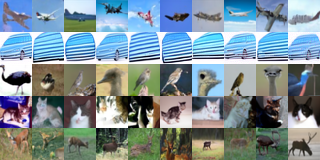}%
  }\hfill
  \subfloat[$\gamma=0.3$]{%
    \includegraphics[width=0.19\textwidth]{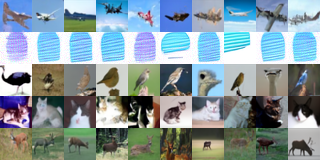}%
  }\hfill
  \subfloat[$\gamma=0.4$]{%
    \includegraphics[width=0.19\textwidth]{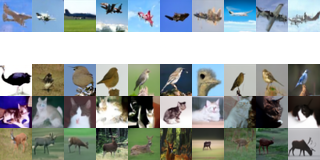}%
  }\hfill
  \subfloat[$\gamma=0.5$]{%
    \includegraphics[width=0.19\textwidth]{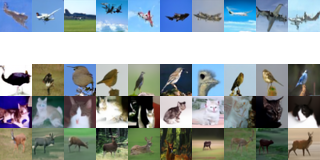}%
  }
  \caption{Generation samples from DDPM models with the class 'automobile' unlearned by CUP. As the unlearning intensity $\gamma$ increases, the shape of the target class becomes progressively vanished.}
  \label{fig:ddpm}
\end{figure*}

Most notably, this strong performance is achieved with significantly greater efficiency. As shown in the last row of Table~\ref{tab:gen}, CUP's Run-Time Efficiency (RTE) is approximately half that of SalUn and more than ten times faster than SA. This computational advantage makes CUP a much more practical and scalable solution for unlearning in large-scale generative models. The qualitative results in Figure~\ref{fig:ddpm} further support these findings, visually confirming that CUP can effectively and controllably remove a target class while preserving the generation quality of the remaining classes. In summary, for image generation, CUP offers a superior trade-off between unlearning performance and computational cost.

\section{Conclusion}

In this work, we addressed the fundamental limitations of the prevailing single-objective optimization (SOO) paradigm in approximate machine unlearning, which often forces an undesirable choice between unlearning efficacy and model fidelity. We argued for and demonstrated the power of reframing unlearning as a Multi-Objective Optimization (MOO) problem. From this new perspective, we developed CUP, a novel algorithm that introduces the Pivoting Gradient Principle to controllably navigate the conflict-free space between two defined anchors: the Efficacy Anchor and the Fidelity Anchor.

Our theoretical analysis guarantees CUP's convergence to a Pareto-stationary point, and our extensive experiments empirically validate its superiority. CUP not only produces solutions closer to the retrained ideal (Optimality, Property 2) but also generates a significantly richer and more diverse set of solutions (Controllability, Property 3), as evidenced by its superior Hypervolume Indicator scores on various vision tasks. Furthermore, this work advocates for the Hypervolume Indicator as a more holistic metric, capable of capturing crucial aspects of utility that are overlooked by traditional, single-outcome evaluations.

In essence, our work reframes unlearning from a single fixed procedure to a dynamic and controllable process. This shift opens up new possibilities for more nuanced, user-centric data governance, such as tiered data removal and fine-grained content moderation. We believe this MOO-based perspective, and the practical control offered by CUP, will serve as a cornerstone for future research in developing more reliable and trustworthy AI systems.




\bibliographystyle{unsrtnat}
\bibliography{reference}

\vfill

\newpage
\appendix

\begin{center}
\Large \textbf{Appendix}
\end{center}

\section{Details for Problem Formulation}\label{app:gen}

\subsection{Image Classification}
For the image classification task, we define the cross-entropy loss as:
\[
l_{\text{CE}}(\rvx, y, \theta) = -\log p_\theta(y | \rvx),
\]
where \((\rvx, y)\) represents the input-label pair  and \(p_\theta(y | \rvx)\) is the predicted probability for \(y\). Based on this, the forgetting loss \(\gL_f\) and remaining loss \(\gL_r\) are defined as:
\begin{align*}
    \gL_f(\theta) = - \frac{1}{|\mathcal{D}_f|} \sum_{(\rvx, y) \in \mathcal{D}_f} l_{\text{CE}}(\rvx, y, \theta), \quad \gL_r(\theta) = \frac{1}{|\mathcal{D}_r|} \sum_{(\rvx, y) \in \mathcal{D}_r} l_{\text{CE}}(\rvx, y, \theta),
\end{align*}
where \(\mathcal{D}_f\) and \(\mathcal{D}_r\) denote the forgetting and remaining datasets, respectively.

\subsection{Image Generation}

For the image generation task, we utilize diffusion models, which are generative frameworks that iteratively transform noise into data through a series of denoising steps.

\paragraph{Diffusion Model}
Diffusion models like DDPM \cite{ho2020ddpm} work by progressively refining random noise to generate data samples. Starting from pure noise, the model learns to reverse a noising process by predicting the added noise at each timestep \( t \). Given a data sample \( \mathbf{x}_0 \) and noise \( \epsilon \sim \mathcal{N}(0, \mathbf{I}) \), a noisy version of the data is created:
\[
\mathbf{x}_t = \sqrt{\bar{\alpha}_t} \, \mathbf{x}_0 + \sqrt{1 - \bar{\alpha}_t} \, \epsilon,
\]
where \( \bar{\alpha}_t \) is the cumulative product of noise schedule parameters controlling the variance at each timestep.

The neural network \( \epsilon_\theta(\mathbf{x}_t, t, c) \) is trained to predict the noise component \( \epsilon \) from the noisy input \( \mathbf{x}_t \), optionally conditioned on information \( c \) (e.g., class labels). To flexibly incorporate conditioning, classifier-free guidance is employed. The final noise prediction \( \hat{\epsilon}_\theta(\mathbf{x}_t | c) \) is computed as:

\[
\hat{\epsilon}_\theta(\mathbf{x}_t | c) = (1 - w) \, \epsilon_\theta(\mathbf{x}_t | \emptyset) + w \, \epsilon_\theta(\mathbf{x}_t | c),
\]

where \( w \in [0, 1] \) is a guidance weight that balances between the conditional and unconditional predictions. Here, \( \epsilon_\theta(\mathbf{x}_t | c) \) represents the noise estimate when conditioning on \( c \), and \( \epsilon_\theta(\mathbf{x}_t | \emptyset) \) is the noise estimate without any conditioning. 

\paragraph{Training and Unlearning Losses}

The diffusion model is trained using a mean-squared error (MSE) loss to minimize the difference between the true noise \( \epsilon \) and the predicted noise \( \epsilon_\theta(\mathbf{x}_t | c) \):

\[
l_{\text{MSE}}(\theta; \mathcal{D}) = \mathbb{E}_{t, \mathbf{x}_0, \epsilon} \left[ \left\| \epsilon - \epsilon_\theta(\mathbf{x}_t | c) \right\|^2 \right],
\]

where \( \mathbf{x}_t \) is defined as above, and \( \epsilon \sim \mathcal{N}(0, \mathbf{I}) \). To adapt the diffusion model for unlearning, we define the forgetting loss \( l_f \) and the remaining loss \( l_r \) based on the MSE loss computed over the forgetting dataset \( \mathcal{D}_f \) and the remaining dataset \( \mathcal{D}_r \):

\begin{align*}
\gL_f(\theta) = -\frac{1}{|\mathcal{D}_f|} \sum_{(\mathbf{x}_0, c) \in \mathcal{D}_f} \mathbb{E}_{t, \epsilon} \left[ \left\| \epsilon - \epsilon_\theta(\mathbf{x}_t | c) \right\|^2 \right], \quad \gL_r(\theta) = \frac{1}{|\mathcal{D}_r|} \sum_{(\mathbf{x}_0, c) \in \mathcal{D}_r} \mathbb{E}_{t, \epsilon} \left[ \left\| \epsilon - \epsilon_\theta(\mathbf{x}_t | c) \right\|^2 \right],
\end{align*}

where \( \mathbf{x}_t \) is generated for each \( \mathbf{x}_0 \) in the respective datasets using the same process.

\section{Experimental Details}\label{app:experiment}

We conduct image classification experiments on Ubuntu 20.04.6 LTS, equipped with an Intel(R) Core(TM) i9-10900X CPU and NVIDIA GeForce RTX 4090 GPU. For all other experiments, we conduct the experiments on Ubuntu 22.04.1 LTS server equipped with AMD Ryzen Threadripper PRO 5975WX, NVIDIA RTX A6000.

\subsection{Details for toy example}\label{app:toy}

For the toy example shown in Figure 1 in our main manuscript, we generate a dataset consisting of 2,000 samples distributed across five distinct Gaussian clusters, each with predefined centers and standard deviations. The cluster centers are located at $[-2, 2]$, $[-6, 6]$, $[5.5, 4]$, $[-4, -4]$, and $[5, -1.0]$, with corresponding standard deviations of 1.5, 1.0, 1.5, 1.5, and 1.5, respectively. For the classification model, we employ a one-layer fully connected neural network with 16 neurons and a ReLU activation function. We firstly train this model for 100 epochs using the ADAM optimizer with a learning rate \(\lambda=10^{-2}\). For unlearning the class 2, we unlearn this model for 20 epochs by GA, WS, CUP. Also, we use \(w_f=w_r=1\) for WS and \(\gamma=0.7\) for CUP.

\subsection{Details for Experimental Settings}\label{app:main}

\paragraph{Image Classification}

For the original model, we train a ResNet-18 for 200 epochs on the CIFAR-10 and SVHN datasets using the SGD optimizer with a cosine-scheduled learning rate initialized at 0.1. We conducted experiments for each class using five different random seeds. Retraining was carried out on the remaining dataset under the same training conditions as the original model. All other unlearning methods were executed for 5 epochs and 20 different hyperparameter settings. Throughout all methods except retraining, data of the same size as the forgetting dataset was randomly sampled from the remaining dataset for unlearning. The detailed hyperparameter configuration for each method is shown in Table \ref{tab:hyperparams_classification}.

\begin{table}[ht]
\centering
\small
\caption{Hyperparameter configurations for each unlearning method in image classification task. All methods utilize exactly 20 settings from their respective hyperparameter spaces.}
\label{tab:hyperparams_classification}
\begin{tabular}{l|l}
\hline
\textbf{Method}       & \textbf{Hyperparameter Space}                                                                                                   \\ \hline
GA, BE, RL            & \(\lambda\) = 20 uniform samples from \([10^{-4}, 10^{-2}]\)                                                                     \\ \hline
BS                    & \(\lambda = [10^{-3}, 5 \times 10^{-3}, 10^{-2}]\), \(\epsilon = [0.01, 0.1, 1, 10]\)                                             \\ \hline
IU                    & \(\alpha = [1, 2, \dots, 20]\)                                                                                                  \\ \hline
\(\ell_1\)-sparse     & \(\lambda = [10^{-3}, 5 \times 10^{-3}, 10^{-2}, 5 \times 10^{-2}, 10^{-1}]\), \(\alpha = [0.001, 0.005, 0.01, 0.05]\)           \\ \hline
SalUn                 & \(\lambda = [10^{-4}, 5 \times 10^{-4}, 10^{-3}, 5 \times 10^{-3}, 10^{-2}]\), \(\textit{threshold} = [10\%, 30\%, 50\%, 70\%]\)      \\ \hline
WS                    & \(\lambda = [10^{-4}, 10^{-3}]\), \(w_f = [10^{-4}, 5 \times 10^{-4}, 10^{-3}, 5 \times 10^{-3}, 10^{-2}, 5 \times 10^{-2}, 0.1, 0.5, 1.0, 5.0]\) \\ \hline
CUP                   & \(\lambda = [10^{-4}, 10^{-3}]\), \(\gamma = [0.01, 0.1, 0.2, 0.3, 0.4, 0.5, 0.6, 0.7, 0.8, 0.9]\)                               \\ \hline
\end{tabular}
\end{table}

\paragraph{Image Generation}

In the image generation task, we employ the DDPM model for sampling, utilizing 1000 time steps on the CIFAR-10 dataset. The pre-trained model was trained for 80,000 epochs using the ADAM optimizer with a learning rate of $\lambda=10^{-4}$. All experiments were conducted with a fixed random seed, and the batch size was set to 128.For SA, we generate 1000 samples for each class to calculate the Fisher Information Matrix (FIM) and unlearn the model for 15,000 epochs with a learning rate of $\lambda=10^{-2}$. The weight of the regularization term $w$ is set to $[10^{-4}, 5 \times 10^{-4}, 10^{-3}, 5 \times 10^{-3}, 10^{-2}, 5 \times 10^{-2}, 0.1, 0.5, 1.0, 5.0]$. For SalUn, we perform 1000 epochs with a learning rate of $\lambda=10^{-3}$ and apply a weight masking threshold of $[10\%, 20\%, 30\%, 40\%, 50\%, 60\%, 70\%, 80\%, 90\%, 100\%]$. For CUP, we unlearn the model for 100 epochs with a learning rate of $\lambda=10^{-3}$, testing 10 different unlearning intensities $\gamma = [0.01, 0.1, 0.2, 0.3, 0.4, 0.5, 0.6, 0.7, 0.8, 0.9]$.

\section{Additional Experimental Results}\label{app:result}

In this section, we present the experimental results of the image classification task for each class and random subset forgetting results.

\begin{table*}[h]
\setlength{\tabcolsep}{2pt}
\scriptsize \centering
\caption{Class-wise unlearning performance in image classification task on CIFAR-10 and SVHN (class 0). The reported values represent the average performance 5 different random seeds. The value within parentheses indicates the standard deviation.}
\resizebox{\textwidth}{!}{
\begin{tabular}{lcccccccccccc}
\toprule
\multirow{2}{*}{\textbf{Methods}} & \multicolumn{6}{c}{\textbf{CIFAR-10}} & \multicolumn{6}{c}{\textbf{SVHN}} \\ 
\cmidrule(lr){2-7} \cmidrule(lr){8-13}
 & RA ($\uparrow$) & UA ($\uparrow$) & TA ($\uparrow$) & MIA ($\uparrow$) & \(\Delta (\downarrow)\) & \(\mathcal{H} (\uparrow)\)  
 & RA ($\uparrow$) & UA ($\uparrow$) & TA ($\uparrow$) & MIA ($\uparrow$) & \(\Delta (\downarrow)\) & \(\mathcal{H} (\uparrow)\)  \\ 
\midrule
Retrain 
& 100.00 (0.00) & 100.00 (0.00) & 94.88 (0.51) & 100.00 (0.00) & \textcolor{blue}{0.00} & \textcolor{blue}{94.88}  
& 100.00 (0.00) & 100.00 (0.00) & 94.85 (0.63) & 100.00 (0.00) & \textcolor{blue}{0.00} & \textcolor{blue}{94.85} \\ 
\midrule

GA 
& 93.45 (0.23) & 94.25 (0.36) & 87.60 (0.15) & 95.64 (0.39) & \textcolor{blue}{12.15} & \textcolor{blue}{76.77}  
& 98.57 (0.27) & 99.13 (0.07) & 92.83 (0.10) & 99.91 (0.01) & \textcolor{blue}{3.41} & \textcolor{blue}{91.38}  \\

WS 
& 92.93 (0.32) & 95.81 (0.62) & 87.37 (0.30) & 96.49 (0.53) & \textcolor{blue}{11.65} & \textcolor{blue}{84.85}  
& 98.53 (0.27) & 100.00 (0.00) & 93.36 (0.09) & 100.00 (0.00) & \textcolor{blue}{2.85} & \textcolor{blue}{94.18}  \\

RL 
& 89.46 (0.67) & 97.78 (0.68) & 84.23 (0.54) & 98.05 (0.64) & \textcolor{blue}{15.25} & \textcolor{blue}{79.80}  
& 97.80 (0.27) & 100.00 (0.00) & 92.56 (0.08) & 100.00 (0.00) & \textcolor{blue}{3.92} & \textcolor{blue}{93.23}  \\

IU 
& 95.64 (2.57) & 74.24 (42.81) & 89.86 (2.84) & 76.36 (40.34) & \textcolor{blue}{35.59} & \textcolor{blue}{58.23}  
& 92.23 (4.82) & 88.47 (19.97) & 88.82 (4.31) & 100.00 (0.00) & \textcolor{blue}{15.56} & \textcolor{blue}{82.99}  \\

BE 
& 99.09 (0.24) & 82.34 (0.32) & 93.48 (0.08) & 98.75 (0.09) & \textcolor{blue}{17.78} & \textcolor{blue}{77.64}  
& 97.35 (0.48) & 100.00 (0.00) & 95.54 (0.02) & 100.00 (0.00) & \textcolor{blue}{3.01} & \textcolor{blue}{94.83}  \\

BS 
& 98.86 (0.39) & 81.39 (12.18) & 93.11 (0.60) & 97.02 (2.15) & \textcolor{blue}{18.96} & \textcolor{blue}{76.09}  
& 99.56 (0.21) & 83.71 (21.47) & 95.34 (0.19) & 100.00 (0.00) & \textcolor{blue}{16.30} & \textcolor{blue}{79.47}  \\

$\ell_{1}$-sparse 
& 88.96 (2.55) & 100.00 (0.00) & 85.50 (2.33) & 100.00 (0.00) & \textcolor{blue}{14.46} & \textcolor{blue}{79.28}  
& 99.95 (0.02) & 98.13 (0.66) & 95.34 (0.06) & 100.00 (0.00) & \textcolor{blue}{1.93} & \textcolor{blue}{95.16}  \\

SalUn 
& 95.52 (0.27) & 94.03 (0.55) & 90.04 (0.14) & 95.60 (0.56) & \textcolor{blue}{9.91} & \textcolor{blue}{87.17}  
& 98.11 (0.15) & 100.00 (0.00) & 92.55 (0.14) & 100.00 (0.00) & \textcolor{blue}{3.76} & \textcolor{blue}{93.21}  \\

CUP 
& 97.39 (0.54) & 97.91 (1.09) & 91.36 (0.60) & 98.53 (0.93) & \textcolor{blue}{\textbf{5.05}} & \textcolor{blue}{\textbf{91.41}}  
& 99.61 (0.22) & 99.99 (0.01) & 95.11 (0.13) & 99.99 (0.01) & \textcolor{blue}{\textbf{0.84}} & \textcolor{blue}{\textbf{95.08}}  \\

\bottomrule
\end{tabular}
}
\end{table*}

\begin{table*}[h]
\setlength{\tabcolsep}{2pt}
\scriptsize \centering
\caption{Class-wise unlearning performance in image classification task on CIFAR-10 and SVHN (class 1). The reported values represent the average performance across 5 different random seeds. The value within parentheses indicates the standard deviation.}
\resizebox{\textwidth}{!}{
\begin{tabular}{lcccccccccccc}
\toprule
\multirow{2}{*}{\textbf{Methods}} & \multicolumn{6}{c}{\textbf{CIFAR-10}} & \multicolumn{6}{c}{\textbf{SVHN}} \\ 
\cmidrule(lr){2-7} \cmidrule(lr){8-13}
 & RA ($\uparrow$) & UA ($\uparrow$) & TA ($\uparrow$) & MIA ($\uparrow$) & \(\Delta (\downarrow)\) & \(\mathcal{H} (\uparrow)\)  
 & RA ($\uparrow$) & UA ($\uparrow$) & TA ($\uparrow$) & MIA ($\uparrow$) & \(\Delta (\downarrow)\) & \(\mathcal{H} (\uparrow)\)  \\ 
\midrule
Retrain 
& 100.00 (0.00) & 100.00 (0.00) & 94.88 (0.51) & 100.00 (0.00) & \textcolor{blue}{0.00} & \textcolor{blue}{94.88}  
& 100.00 (0.00) & 100.00 (0.00) & 94.85 (0.63) & 100.00 (0.00) & \textcolor{blue}{0.00} & \textcolor{blue}{94.85} \\ 
\midrule

GA 
& 95.75 (0.31) & 95.77 (0.26) & 89.20 (0.10) & 97.99 (0.18) & \textcolor{blue}{8.27} & \textcolor{blue}{83.50}  
& 88.37 (0.23) & 83.53 (0.33) & 94.92 (0.05) & 19.15 (0.28) & \textcolor{blue}{83.33} & \textcolor{blue}{13.94}  \\

WS
& 96.03 (0.30) & 97.80 (0.50) & 89.58 (0.20) & 98.09 (0.44) & \textcolor{blue}{6.98} & \textcolor{blue}{88.07}  
& 94.65 (0.16) & 42.09 (0.55) & 95.53 (0.06) & 52.38 (0.88) & \textcolor{blue}{75.16} & \textcolor{blue}{27.09}  \\

RL 
& 93.54 (0.42) & 98.49 (0.28) & 87.28 (0.32) & 98.65 (0.26) & \textcolor{blue}{9.91} & \textcolor{blue}{84.17}  
& 94.65 (0.16) & 42.09 (0.55) & 95.53 (0.06) & 52.38 (0.88) & \textcolor{blue}{75.16} & \textcolor{blue}{25.44}  \\

IU 
& 90.55 (6.78) & 75.08 (43.17) & 84.96 (6.63) & 76.98 (39.88) & \textcolor{blue}{36.50} & \textcolor{blue}{52.98}  
& 93.74 (3.76) & 44.60 (27.09) & 95.08 (0.39) & 25.02 (14.85) & \textcolor{blue}{93.44} & \textcolor{blue}{17.02}  \\

BE 
& 98.62 (0.29) & 71.16 (0.45) & 92.13 (0.10) & 99.87 (0.05) & \textcolor{blue}{28.97} & \textcolor{blue}{67.18}  
& 84.70 (0.59) & 100.00 (0.00) & 87.43 (0.58) & 0.00 (0.01) & \textcolor{blue}{101.48} & \textcolor{blue}{1.16}  \\

BS 
& 99.41 (0.24) & 47.74 (0.63) & 93.48 (0.05) & 99.40 (0.24) & \textcolor{blue}{52.28} & \textcolor{blue}{44.22}  
& 72.65 (18.20) & 76.23 (31.61) & 72.73 (17.93) & 31.26 (31.27) & \textcolor{blue}{80.95} & \textcolor{blue}{19.50}  \\

$\ell_{1}$-sparse 
& 89.65 (0.89) & 100.00 (0.00) & 85.75 (1.10) & 100.00 (0.00) & \textcolor{blue}{13.57} & \textcolor{blue}{80.25}  
& 76.02 (6.99) & 35.40 (18.14) & 76.09 (5.63) & 35.56 (9.78) & \textcolor{blue}{96.31} & \textcolor{blue}{11.28}  \\

SalUn 
& 97.43 (0.26) & 98.71 (0.81) & 91.08 (0.21) & 99.17 (0.65) & \textcolor{blue}{4.56} & \textcolor{blue}{89.51}  
& 94.81 (0.18) & 40.30 (0.29) & 95.52 (0.06) & 54.28 (0.99) & \textcolor{blue}{75.38} & \textcolor{blue}{30.65}  \\

CUP 
& 98.68 (0.38) & 98.83 (0.40) & 92.19 (0.19) & 99.56 (0.21) & \textcolor{blue}{\textbf{2.95}} & \textcolor{blue}{\textbf{91.92}}  
& 94.87 (0.38) & 41.01 (1.52) & 95.55 (0.07) & 53.75 (2.12) & \textcolor{blue}{\textbf{75.13}} & \textcolor{blue}{\textbf{29.49}} \\

\bottomrule
\end{tabular}
}
\end{table*}

\begin{table*}[h]
\setlength{\tabcolsep}{2pt}
\scriptsize \centering
\caption{Class-wise unlearning performance in image classification task on CIFAR-10 and SVHN (class 2). The reported values represent the average performance across 5 different random seeds. The value within parentheses indicates the standard deviation.}
\resizebox{\textwidth}{!}{
\begin{tabular}{lcccccccccccc}
\toprule
\multirow{2}{*}{\textbf{Methods}} & \multicolumn{6}{c}{\textbf{CIFAR-10}} & \multicolumn{6}{c}{\textbf{SVHN}} \\ 
\cmidrule(lr){2-7} \cmidrule(lr){8-13}
 & RA ($\uparrow$) & UA ($\uparrow$) & TA ($\uparrow$) & MIA ($\uparrow$) & \(\Delta (\downarrow)\) & \(\mathcal{H} (\uparrow)\)  
 & RA ($\uparrow$) & UA ($\uparrow$) & TA ($\uparrow$) & MIA ($\uparrow$) & \(\mathcal{H} (\uparrow)\) & \(\Delta (\downarrow)\) \\ 
\midrule
Retrain 
& 100.00 (0.00) & 100.00 (0.00) & 94.88 (0.51) & 100.00 (0.00) & \textcolor{blue}{0.00} & \textcolor{blue}{94.88}  
& 100.00 (0.00) & 100.00 (0.00) & 94.85 (0.63) & 100.00 (0.00) & \textcolor{blue}{94.85} & \textcolor{blue}{0.00} \\ 
\midrule

GA 
& 93.09 (0.29) & 91.28 (0.24) & 87.24 (0.07) & 92.98 (0.35) & \textcolor{blue}{15.29} & \textcolor{blue}{76.90}  
& 91.43 (0.15) & 92.88 (0.46) & 94.30 (0.14) & 10.18 (0.19) & \textcolor{blue}{90.52} & \textcolor{blue}{8.21}  \\

WS
& 94.93 (0.57) & 98.38 (0.58) & 89.58 (0.39) & 98.66 (0.53) & \textcolor{blue}{7.74} & \textcolor{blue}{87.98}  
& 96.08 (0.24) & 46.39 (2.07) & 94.92 (0.12) & 27.74 (4.60) & \textcolor{blue}{90.07} & \textcolor{blue}{16.63}  \\

RL 
& 91.91 (0.68) & 99.71 (0.20) & 87.12 (0.50) & 99.78 (0.18) & \textcolor{blue}{11.33} & \textcolor{blue}{84.03}  
& 96.08 (0.24) & 46.39 (2.07) & 94.92 (0.12) & 27.74 (4.60) & \textcolor{blue}{90.07} & \textcolor{blue}{16.03}  \\

IU 
& 91.96 (7.07) & 74.83 (42.78) & 87.12 (6.82) & 78.37 (37.29) & \textcolor{blue}{35.05} & \textcolor{blue}{54.51}  
& 97.12 (1.67) & 30.37 (17.63) & 94.86 (0.50) & 60.92 (33.91) & \textcolor{blue}{79.91} & \textcolor{blue}{22.12}  \\

BE 
& 99.15 (0.23) & 78.56 (0.58) & 93.41 (0.05) & 98.19 (0.08) & \textcolor{blue}{21.59} & \textcolor{blue}{75.30}  
& 92.70 (0.20) & 83.18 (1.08) & 95.36 (0.05) & 2.68 (0.66) & \textcolor{blue}{99.03} & \textcolor{blue}{6.55}  \\

BS 
& 98.93 (0.54) & 77.11 (13.47) & 93.24 (0.94) & 96.57 (2.21) & \textcolor{blue}{23.24} & \textcolor{blue}{71.88}  
& 75.43 (17.47) & 61.47 (27.80) & 73.03 (17.27) & 35.74 (26.02) & \textcolor{blue}{82.00} & \textcolor{blue}{19.17}  \\

$\ell_{1}$-sparse 
& 90.18 (0.77) & 100.00 (0.00) & 86.90 (0.72) & 100.00 (0.00) & \textcolor{blue}{12.75} & \textcolor{blue}{79.51}  
& 76.17 (2.46) & 21.40 (11.98) & 74.11 (4.18) & 42.14 (16.03) & \textcolor{blue}{102.73} & \textcolor{blue}{10.13}  \\

SalUn 
& 96.60 (0.30) & 97.26 (0.32) & 91.13 (0.07) & 97.98 (0.29) & \textcolor{blue}{6.20} & \textcolor{blue}{89.55}  
& 90.75 (0.22) & 92.61 (0.86) & 93.77 (0.12) & 11.15 (1.12) & \textcolor{blue}{89.65} & \textcolor{blue}{16.37}  \\

CUP 
& 97.81 (0.45) & 99.63 (0.15) & 91.87 (0.27) & 99.79 (0.09) & \textcolor{blue}{\textbf{3.87}} & \textcolor{blue}{\textbf{92.15}}  
& 95.24 (1.06) & 72.60 (9.18) & 95.00 (0.16) & 46.28 (21.84) & \textcolor{blue}{\textbf{60.50}} & \textcolor{blue}{\textbf{34.52}}  \\

\bottomrule
\end{tabular}
}
\end{table*}

\begin{table*}[h]
\setlength{\tabcolsep}{2pt}
\scriptsize \centering
\caption{Class-wise unlearning performance in image classification task on CIFAR-10 and SVHN (class 3). The reported values represent the average performance across 5 different random seeds. The value within parentheses indicates the standard deviation.}
\resizebox{\textwidth}{!}{
\begin{tabular}{lcccccccccccc}
\toprule
\multirow{2}{*}{\textbf{Methods}} & \multicolumn{6}{c}{\textbf{CIFAR-10}} & \multicolumn{6}{c}{\textbf{SVHN}} \\ 
\cmidrule(lr){2-7} \cmidrule(lr){8-13}
 & RA ($\uparrow$) & UA ($\uparrow$) & TA ($\uparrow$) & MIA ($\uparrow$) & \(\Delta (\downarrow)\) & \(\mathcal{H} (\uparrow)\)  
 & RA ($\uparrow$) & UA ($\uparrow$) & TA ($\uparrow$) & MIA ($\uparrow$) & \(\Delta (\downarrow)\) & \(\mathcal{H} (\uparrow)\)  \\ 
\midrule
Retrain 
& 100.00 (0.00) & 100.00 (0.00) & 94.88 (0.51) & 100.00 (0.00) & \textcolor{blue}{0.00} & \textcolor{blue}{94.88}  
& 100.00 (0.00) & 100.00 (0.00) & 94.85 (0.63) & 100.00 (0.00) & \textcolor{blue}{0.00} & \textcolor{blue}{94.85} \\ 
\midrule

GA 
& 96.68 (0.20) & 93.72 (0.18) & 91.52 (0.03) & 96.32 (0.19) & \textcolor{blue}{9.19} & \textcolor{blue}{85.02}  
& 93.41 (0.20) & 96.29 (0.11) & 93.68 (0.11) & 5.45 (0.28) & \textcolor{blue}{94.88} & \textcolor{blue}{4.59} \\

WS 
& 95.42 (0.27) & 97.10 (0.46) & 90.85 (0.23) & 97.69 (0.41) & \textcolor{blue}{7.85} & \textcolor{blue}{89.99}  
& 96.42 (0.24) & 60.57 (1.34) & 95.11 (0.10) & 86.11 (0.59) & \textcolor{blue}{41.97} & \textcolor{blue}{49.91}  \\

RL 
& 93.01 (0.33) & 98.54 (0.25) & 88.64 (0.29) & 98.76 (0.16) & \textcolor{blue}{10.36} & \textcolor{blue}{86.20}  
& 96.42 (0.24) & 60.57 (1.34) & 95.11 (0.10) & 86.11 (0.59) & \textcolor{blue}{41.97} & \textcolor{blue}{51.40} \\

IU 
& 98.71 (0.77) & 73.79 (42.60) & 94.06 (0.98) & 75.32 (41.84) & \textcolor{blue}{36.08} & \textcolor{blue}{54.71}  
& 97.60 (1.39) & 40.31 (23.39) & 95.97 (0.19) & 72.96 (40.78) & \textcolor{blue}{65.57} & \textcolor{blue}{41.99}  \\

BE 
& 99.56 (0.18) & 70.62 (0.18) & 95.15 (0.09) & 97.95 (0.21) & \textcolor{blue}{29.47} & \textcolor{blue}{77.11}  
& 31.11 (5.15) & 100.00 (0.00) & 27.13 (4.40) & 100.00 (0.00) & \textcolor{blue}{97.49} & \textcolor{blue}{9.87} \\

BS 
& 99.50 (0.23) & 76.04 (6.28) & 95.05 (0.57) & 97.52 (0.75) & \textcolor{blue}{24.12} & \textcolor{blue}{76.76}  
& 89.43 (6.48) & 48.27 (12.44) & 83.94 (8.54) & 64.13 (23.24) & \textcolor{blue}{64.99} & \textcolor{blue}{29.32}  \\

$\ell_{1}$-sparse 
& 91.06 (0.85) & 100.00 (0.00) & 87.99 (1.01) & 100.00 (0.00) & \textcolor{blue}{12.03} & \textcolor{blue}{85.68}  
& 60.69 (12.61) & 83.75 (13.27) & 61.97 (11.45) & 42.71 (45.64) & \textcolor{blue}{79.10} & \textcolor{blue}{24.50}  \\

SalUn 
& 97.32 (0.27) & 97.41 (0.48) & 92.76 (0.26) & 98.50 (0.34) & \textcolor{blue}{5.19} & \textcolor{blue}{91.53}  
& 96.52 (0.24) & 58.79 (1.24) & 95.15 (0.08) & 85.16 (0.61) & \textcolor{blue}{43.95} & \textcolor{blue}{47.61}  \\

CUP 
& 98.47 (0.21) & 98.87 (0.61) & 93.56 (0.17) & 99.29 (0.43) & \textcolor{blue}{\textbf{3.21}} & \textcolor{blue}{\textbf{93.87}}  
& 96.19 (0.45) & 84.12 (3.23) & 95.81 (0.10) & 70.87 (35.28) & \textcolor{blue}{\textbf{33.39}} & \textcolor{blue}{\textbf{70.46}}  \\

\bottomrule
\end{tabular}
}
\end{table*}

\begin{table*}[h]
\setlength{\tabcolsep}{2pt}
\scriptsize \centering
\caption{Class-wise unlearning performance in image classification task on CIFAR-10 and SVHN (class 4). The reported values represent the average performance across 5 different random seeds. The value within parentheses indicates the standard deviation.}
\resizebox{\textwidth}{!}{
\begin{tabular}{lcccccccccccc}
\toprule
\multirow{2}{*}{\textbf{Methods}} & \multicolumn{6}{c}{\textbf{CIFAR-10}} & \multicolumn{6}{c}{\textbf{SVHN}} \\ 
\cmidrule(lr){2-7} \cmidrule(lr){8-13}
 & RA ($\uparrow$) & UA ($\uparrow$) & TA ($\uparrow$) & MIA ($\uparrow$) & \(\Delta (\downarrow)\) & \(\mathcal{H} (\uparrow)\)  
 & RA ($\uparrow$) & UA ($\uparrow$) & TA ($\uparrow$) & MIA ($\uparrow$) & \(\Delta (\downarrow)\) & \(\mathcal{H} (\uparrow)\)  \\ 
\midrule
Retrain 
& 100.00 (0.00) & 100.00 (0.00) & 94.88 (0.51) & 100.00 (0.00) & \textcolor{blue}{0.00} & \textcolor{blue}{94.88}  
& 100.00 (0.00) & 100.00 (0.00) & 94.85 (0.63) & 100.00 (0.00) & \textcolor{blue}{0.00} & \textcolor{blue}{94.85} \\ 
\midrule

GA 
& 93.14 (0.32) & 94.31 (0.19) & 86.58 (0.23) & 97.11 (0.08) & \textcolor{blue}{12.32} & \textcolor{blue}{76.55}  
& 95.80 (0.20) & 95.72 (0.18) & 94.20 (0.14) & 99.25 (0.05) & \textcolor{blue}{\textbf{6.24}} & \textcolor{blue}{\textbf{90.01}}  \\

WS
& 93.92 (0.36) & 99.97 (0.02) & 88.04 (0.33) & 99.99 (0.01) & \textcolor{blue}{8.94} & \textcolor{blue}{86.56}  
& 97.50 (0.23) & 62.65 (1.08) & 95.39 (0.03) & 86.94 (0.76) & \textcolor{blue}{39.65} & \textcolor{blue}{59.18}  \\

RL 
& 89.88 (0.49) & 100.00 (0.00) & 84.62 (0.52) & 100.00 (0.00) & \textcolor{blue}{14.21} & \textcolor{blue}{80.43}  
& 97.50 (0.23) & 62.65 (1.08) & 95.39 (0.03) & 86.94 (0.76) & \textcolor{blue}{39.65} & \textcolor{blue}{58.84} \\

IU 
& 88.54 (7.05) & 74.99 (43.11) & 83.49 (6.77) & 77.77 (38.26) & \textcolor{blue}{37.06} & \textcolor{blue}{55.81}  
& 93.29 (4.11) & 76.72 (40.33) & 91.21 (3.07) & 99.89 (0.19) & \textcolor{blue}{24.65} & \textcolor{blue}{70.25}  \\

BE 
& 98.93 (0.21) & 82.81 (0.66) & 92.76 (0.03) & 99.85 (0.03) & \textcolor{blue}{17.32} & \textcolor{blue}{81.50}  
& 96.18 (0.17) & 94.00 (0.32) & 95.64 (0.01) & 54.04 (27.90) & \textcolor{blue}{46.51} & \textcolor{blue}{48.87} \\

BS 
& 98.57 (0.69) & 83.53 (13.21) & 92.15 (1.39) & 99.61 (0.30) & \textcolor{blue}{16.72} & \textcolor{blue}{77.47}  
& 96.73 (1.14) & 74.83 (32.62) & 94.80 (0.54) & 83.92 (22.90) & \textcolor{blue}{30.06} & \textcolor{blue}{58.96}  \\

$\ell_{1}$-sparse 
& 90.36 (1.41) & 100.00 (0.00) & 86.59 (1.14) & 100.00 (0.00) & \textcolor{blue}{12.53} & \textcolor{blue}{80.89}  
& 72.18 (9.89) & 56.58 (22.92) & 72.35 (9.06) & 56.18 (23.56) & \textcolor{blue}{71.60} & \textcolor{blue}{23.65}  \\

SalUn 
& 96.27 (0.28) & 99.24 (0.16) & 90.33 (0.17) & 99.70 (0.04) & \textcolor{blue}{5.72} & \textcolor{blue}{88.61}  
& 97.57 (0.23) & 60.58 (1.20) & 95.44 (0.03) & 86.23 (0.78) & \textcolor{blue}{41.83} & \textcolor{blue}{56.04}  \\

CUP 
& 98.37 (0.31) & 98.95 (0.40) & 92.18 (0.33) & 99.69 (0.18) & \textcolor{blue}{\textbf{3.11}} & \textcolor{blue}{\textbf{92.23}}  
& 97.01 (0.45) & 86.01 (3.74) & 95.25 (0.16) & 96.40 (1.26) & \textcolor{blue}{14.76} & \textcolor{blue}{86.21}  \\

\bottomrule
\end{tabular}
}
\end{table*}

\begin{table*}[h]
\setlength{\tabcolsep}{2pt}
\scriptsize \centering
\caption{Class-wise unlearning performance in image classification task on CIFAR-10 and SVHN (class 5). The reported values represent the average performance across 5 different random seeds. The value within parentheses indicates the standard deviation.}
\resizebox{\textwidth}{!}{
\begin{tabular}{lcccccccccccc}
\toprule
\multirow{2}{*}{\textbf{Methods}} & \multicolumn{6}{c}{\textbf{CIFAR-10}} & \multicolumn{6}{c}{\textbf{SVHN}} \\ 
\cmidrule(lr){2-7} \cmidrule(lr){8-13}
 & RA ($\uparrow$) & UA ($\uparrow$) & TA ($\uparrow$) & MIA ($\uparrow$) & \(\Delta (\downarrow)\) & \(\mathcal{H} (\uparrow)\)  
 & RA ($\uparrow$) & UA ($\uparrow$) & TA ($\uparrow$) & MIA ($\uparrow$) & \(\Delta (\downarrow)\) & \(\mathcal{H} (\uparrow)\)  \\ 
\midrule
Retrain 
& 100.00 (0.00) & 100.00 (0.00) & 94.88 (0.51) & 100.00 (0.00) & \textcolor{blue}{0.00} & \textcolor{blue}{94.88}  
& 100.00 (0.00) & 100.00 (0.00) & 94.85 (0.63) & 100.00 (0.00) & \textcolor{blue}{0.00} & \textcolor{blue}{94.85} \\ 
\midrule

GA 
& 92.21 (0.18) & 94.71 (0.19) & 87.04 (0.13) & 96.99 (0.13) & \textcolor{blue}{13.05} & \textcolor{blue}{77.80}  
& 96.55 (0.20) & 96.87 (0.28) & 94.04 (0.07) & 99.44 (0.06) & \textcolor{blue}{5.05} & \textcolor{blue}{90.49}  \\

WS 
& 94.31 (0.34) & 97.18 (0.31) & 89.18 (0.33) & 97.69 (0.25) & \textcolor{blue}{9.29} & \textcolor{blue}{87.49}  
& 96.47 (0.19) & 98.30 (0.18) & 94.67 (0.11) & 99.21 (0.22) & \textcolor{blue}{4.18} & \textcolor{blue}{92.45}  \\

RL 
& 91.64 (0.55) & 98.56 (0.33) & 87.07 (0.44) & 98.72 (0.31) & \textcolor{blue}{12.07} & \textcolor{blue}{83.53}  
& 95.99 (0.25) & 99.50 (0.17) & 94.13 (0.19) & 99.68 (0.17) & \textcolor{blue}{4.42} & \textcolor{blue}{91.88}  \\

IU 
& 98.59 (0.88) & 74.23 (42.80) & 93.67 (1.12) & 75.88 (41.12) & \textcolor{blue}{35.37} & \textcolor{blue}{59.44}  
& 95.37 (2.89) & 75.06 (43.19) & 91.27 (3.08) & 96.04 (6.86) & \textcolor{blue}{26.09} & \textcolor{blue}{76.71}  \\

BE 
& 99.16 (0.23) & 74.23 (0.48) & 93.87 (0.09) & 98.83 (0.05) & \textcolor{blue}{25.86} & \textcolor{blue}{78.85}  
& 96.94 (0.17) & 95.00 (0.55) & 95.51 (0.04) & 100.00 (0.00) & \textcolor{blue}{5.87} & \textcolor{blue}{91.97}  \\

BS 
& 99.15 (0.29) & 80.10 (10.28) & 93.84 (0.56) & 98.44 (0.75) & \textcolor{blue}{20.06} & \textcolor{blue}{78.49}  
& 97.37 (0.81) & 77.41 (30.26) & 95.21 (0.08) & 96.75 (4.70) & \textcolor{blue}{22.98} & \textcolor{blue}{72.33}  \\

$\ell_{1}$-sparse 
& 90.69 (1.43) & 100.00 (0.00) & 87.28 (1.06) & 100.00 (0.00) & \textcolor{blue}{12.45} & \textcolor{blue}{81.52}  
& 79.25 (4.14) & 56.82 (17.71) & 78.42 (5.03) & 68.52 (24.12) & \textcolor{blue}{59.93} & \textcolor{blue}{36.10}  \\

SalUn 
& 96.27 (0.33) & 97.11 (0.70) & 90.99 (0.26) & 98.09 (0.52) & \textcolor{blue}{6.83} & \textcolor{blue}{89.20}  
& 96.13 (0.20) & 99.23 (0.20) & 94.23 (0.15) & 99.65 (0.19) & \textcolor{blue}{4.30} & \textcolor{blue}{92.10}  \\

CUP 
& 97.68 (0.39) & 98.90 (0.44) & 92.17 (0.31) & 99.29 (0.36) & \textcolor{blue}{\textbf{4.31}} & \textcolor{blue}{\textbf{92.09}}  
& 96.86 (0.25) & 98.78 (0.30) & 95.08 (0.17) & 99.64 (0.13) & \textcolor{blue}{\textbf{3.48}} & \textcolor{blue}{\textbf{93.50}}  \\

\bottomrule
\end{tabular}
}
\end{table*}

\begin{table*}[h]
\setlength{\tabcolsep}{2pt}
\scriptsize \centering
\caption{Class-wise unlearning performance in image classification task on CIFAR-10 and SVHN (class 6). The reported values represent the average performance across 5 different random seeds. The value within parentheses indicates the standard deviation.}
\resizebox{\textwidth}{!}{
\begin{tabular}{lcccccccccccc}
\toprule
\multirow{2}{*}{\textbf{Methods}} & \multicolumn{6}{c}{\textbf{CIFAR-10}} & \multicolumn{6}{c}{\textbf{SVHN}} \\ 
\cmidrule(lr){2-7} \cmidrule(lr){8-13}
 & RA ($\uparrow$) & UA ($\uparrow$) & TA ($\uparrow$) & MIA ($\uparrow$) & \(\Delta (\downarrow)\) & \(\mathcal{H} (\uparrow)\)  
 & RA ($\uparrow$) & UA ($\uparrow$) & TA ($\uparrow$) & MIA ($\uparrow$) & \(\Delta (\downarrow)\) & \(\mathcal{H} (\uparrow)\)  \\ 
\midrule
Retrain 
& 100.00 (0.00) & 100.00 (0.00) & 94.88 (0.51) & 100.00 (0.00) & \textcolor{blue}{0.00} & \textcolor{blue}{94.88}  
& 100.00 (0.00) & 100.00 (0.00) & 94.85 (0.63) & 100.00 (0.00) & \textcolor{blue}{0.00} & \textcolor{blue}{94.85} \\ 
\midrule

GA 
& 93.81 (0.19) & 95.72 (0.16) & 87.24 (0.13) & 97.58 (0.16) & \textcolor{blue}{10.65} & \textcolor{blue}{77.75}  
& 97.59 (0.23) & 97.52 (0.10) & 93.51 (0.04) & 99.63 (0.04) & \textcolor{blue}{4.23} & \textcolor{blue}{90.80}  \\

WS 
& 93.09 (0.70) & 96.19 (0.96) & 87.16 (0.45) & 96.90 (0.74) & \textcolor{blue}{11.14} & \textcolor{blue}{83.97}  
& 97.57 (0.23) & 99.69 (0.03) & 94.15 (0.10) & 99.88 (0.04) & \textcolor{blue}{3.03} & \textcolor{blue}{93.35}  \\

RL 
& 88.88 (0.68) & 98.39 (0.34) & 83.61 (0.53) & 98.62 (0.35) & \textcolor{blue}{15.63} & \textcolor{blue}{77.84}  
& 96.58 (0.26) & 99.93 (0.02) & 93.37 (0.11) & 99.96 (0.02) & \textcolor{blue}{4.27} & \textcolor{blue}{92.26}  \\

IU 
& 92.74 (4.70) & 74.88 (43.16) & 86.85 (4.77) & 76.56 (40.47) & \textcolor{blue}{35.92} & \textcolor{blue}{56.82}  
& 94.87 (3.00) & 75.00 (43.30) & 91.13 (2.77) & 95.39 (7.99) & \textcolor{blue}{26.37} & \textcolor{blue}{70.98}  \\

BE 
& 98.42 (0.25) & 85.59 (0.76) & 91.83 (0.04) & 99.62 (0.03) & \textcolor{blue}{14.73} & \textcolor{blue}{77.89}  
& 98.55 (0.17) & 96.90 (0.56) & 95.64 (0.03) & 100.00 (0.00) & \textcolor{blue}{3.44} & \textcolor{blue}{93.89}  \\

BS 
& 98.43 (0.75) & 81.68 (16.15) & 91.82 (1.40) & 99.21 (0.33) & \textcolor{blue}{18.58} & \textcolor{blue}{75.04}  
& 98.71 (0.28) & 80.18 (23.65) & 95.44 (0.08) & 99.98 (0.04) & \textcolor{blue}{19.87} & \textcolor{blue}{75.67}  \\

$\ell_{1}$-sparse 
& 89.81 (1.12) & 100.00 (0.00) & 85.99 (1.16) & 100.00 (0.00) & \textcolor{blue}{13.20} & \textcolor{blue}{78.51}  
& 84.92 (1.82) & 100.00 (0.00) & 84.66 (1.82) & 60.00 (48.99) & \textcolor{blue}{44.21} & \textcolor{blue}{59.90}  \\

SalUn 
& 93.92 (0.55) & 94.69 (1.10) & 88.16 (0.35) & 96.13 (0.80) & \textcolor{blue}{10.90} & \textcolor{blue}{83.96}  
& 96.54 (0.25) & 99.87 (0.02) & 93.35 (0.07) & 99.94 (0.02) & \textcolor{blue}{4.31} & \textcolor{blue}{92.46}  \\

CUP 
& 96.32 (0.75) & 97.71 (0.53) & 89.89 (0.55) & 98.22 (0.41) & \textcolor{blue}{\textbf{6.49}} & \textcolor{blue}{\textbf{90.50}}  
& 98.43 (0.24) & 99.28 (0.07) & 95.08 (0.13) & 99.79 (0.05) & \textcolor{blue}{\textbf{1.93}} & \textcolor{blue}{\textbf{94.46}}  \\

\bottomrule
\end{tabular}
}
\end{table*}

\begin{table*}[h]
\setlength{\tabcolsep}{2pt}
\scriptsize \centering
\caption{Class-wise unlearning performance in image classification task on CIFAR-10 and SVHN (class 7). The reported values represent the average performance across 5 different random seeds. The value within parentheses indicates the standard deviation.}
\resizebox{\textwidth}{!}{
\begin{tabular}{lcccccccccccc}
\toprule
\multirow{2}{*}{\textbf{Methods}} & \multicolumn{6}{c}{\textbf{CIFAR-10}} & \multicolumn{6}{c}{\textbf{SVHN}} \\ 
\cmidrule(lr){2-7} \cmidrule(lr){8-13}
 & RA ($\uparrow$) & UA ($\uparrow$) & TA ($\uparrow$) & MIA ($\uparrow$) & \(\Delta (\downarrow)\) & \(\mathcal{H} (\uparrow)\)  
 & RA ($\uparrow$) & UA ($\uparrow$) & TA ($\uparrow$) & MIA ($\uparrow$) & \(\Delta (\downarrow)\) & \(\mathcal{H} (\uparrow)\)  \\ 
\midrule
Retrain 
& 100.00 (0.00) & 100.00 (0.00) & 94.88 (0.51) & 100.00 (0.00) & \textcolor{blue}{0.00} & \textcolor{blue}{94.88}  
& 100.00 (0.00) & 100.00 (0.00) & 94.85 (0.63) & 100.00 (0.00) & \textcolor{blue}{0.00} & \textcolor{blue}{94.85} \\ 
\midrule

GA 
& 96.26 (0.33) & 92.96 (0.17) & 89.52 (0.11) & 96.74 (0.15) & \textcolor{blue}{9.90} & \textcolor{blue}{82.57}  
& 98.35 (0.19) & 97.55 (0.20) & 94.45 (0.11) & 99.47 (0.13) & \textcolor{blue}{3.29} & \textcolor{blue}{91.86}  \\

WS 
& 95.08 (0.44) & 98.42 (0.22) & 89.10 (0.27) & 98.64 (0.21) & \textcolor{blue}{7.52} & \textcolor{blue}{87.82}  
& 98.14 (0.11) & 98.73 (0.18) & 94.83 (0.13) & 98.95 (0.14) & \textcolor{blue}{2.66} & \textcolor{blue}{93.61}  \\

RL 
& 92.41 (0.52) & 99.06 (0.30) & 86.76 (0.37) & 99.15 (0.28) & \textcolor{blue}{10.84} & \textcolor{blue}{83.48}  
& 97.65 (0.12) & 99.13 (0.14) & 94.38 (0.11) & 99.20 (0.11) & \textcolor{blue}{2.99} & \textcolor{blue}{93.75}  \\

IU 
& 80.64 (13.99) & 74.54 (42.99) & 76.41 (13.09) & 76.20 (40.76) & \textcolor{blue}{43.73} & \textcolor{blue}{47.52}  
& 96.86 (2.34) & 75.00 (43.30) & 92.88 (2.14) & 94.54 (9.46) & \textcolor{blue}{25.95} & \textcolor{blue}{74.56}  \\

BE 
& 99.30 (0.24) & 73.96 (0.37) & 93.36 (0.06) & 99.81 (0.02) & \textcolor{blue}{26.07} & \textcolor{blue}{77.03}  
& 98.74 (0.17) & 96.01 (0.49) & 95.46 (0.02) & 100.00 (0.00) & \textcolor{blue}{4.20} & \textcolor{blue}{93.41}  \\

BS 
& 99.21 (0.32) & 78.83 (15.34) & 93.16 (0.49) & 99.52 (0.49) & \textcolor{blue}{21.22} & \textcolor{blue}{74.23}  
& 98.89 (0.27) & 80.33 (22.25) & 95.51 (0.05) & 99.93 (0.10) & \textcolor{blue}{19.71} & \textcolor{blue}{76.01}  \\

$\ell_{1}$-sparse 
& 87.66 (1.41) & 100.00 (0.00) & 84.37 (1.15) & 100.00 (0.00) & \textcolor{blue}{15.90} & \textcolor{blue}{75.78}  
& 83.87 (1.07) & 84.52 (20.16) & 82.92 (1.60) & 83.96 (24.14) & \textcolor{blue}{30.38} & \textcolor{blue}{64.13}  \\

SalUn 
& 96.64 (0.32) & 98.88 (0.24) & 90.72 (0.22) & 99.30 (0.17) & \textcolor{blue}{5.15} & \textcolor{blue}{88.99}  
& 97.92 (0.15) & 99.25 (0.17) & 94.55 (0.12) & 99.52 (0.18) & \textcolor{blue}{2.59} & \textcolor{blue}{93.94}  \\

CUP 
& 97.86 (0.44) & 98.32 (0.18) & 91.61 (0.31) & 98.62 (0.20) & \textcolor{blue}{\textbf{4.13}} & \textcolor{blue}{\textbf{91.68}}  
& 98.37 (0.16) & 98.98 (0.14) & 95.01 (0.12) & 99.05 (0.12) & \textcolor{blue}{\textbf{2.29}} & \textcolor{blue}{\textbf{94.05}}  \\

\bottomrule
\end{tabular}
}
\end{table*}

\begin{table*}[h]
\setlength{\tabcolsep}{2pt}
\scriptsize \centering
\caption{Class-wise unlearning performance in image classification task on CIFAR-10 and SVHN (class 8). The reported values represent the average performance across 5 different random seeds. The value within parentheses indicates the standard deviation.}
\resizebox{\textwidth}{!}{
\begin{tabular}{lcccccccccccc}
\toprule
\multirow{2}{*}{\textbf{Methods}} & \multicolumn{6}{c}{\textbf{CIFAR-10}} & \multicolumn{6}{c}{\textbf{SVHN}} \\ 
\cmidrule(lr){2-7} \cmidrule(lr){8-13}
 & RA ($\uparrow$) & UA ($\uparrow$) & TA ($\uparrow$) & MIA ($\uparrow$) & \(\Delta (\downarrow)\) & \(\mathcal{H} (\uparrow)\)  
 & RA ($\uparrow$) & UA ($\uparrow$) & TA ($\uparrow$) & MIA ($\uparrow$) & \(\Delta (\downarrow)\) & \(\mathcal{H} (\uparrow)\)  \\ 
\midrule
Retrain 
& 100.00 (0.00) & 100.00 (0.00) & 94.88 (0.51) & 100.00 (0.00) & \textcolor{blue}{0.00} & \textcolor{blue}{94.88}  
& 100.00 (0.00) & 100.00 (0.00) & 94.85 (0.63) & 100.00 (0.00) & \textcolor{blue}{0.00} & \textcolor{blue}{94.85} \\ 
\midrule

GA 
& 92.45 (0.19) & 94.87 (0.27) & 86.39 (0.14) & 96.72 (0.23) & \textcolor{blue}{12.76} & \textcolor{blue}{76.83}  
& 98.97 (0.22) & 97.99 (0.10) & 93.88 (0.05) & 99.77 (0.04) & \textcolor{blue}{3.00} & \textcolor{blue}{92.76}  \\

WS 
& 94.65 (0.65) & 95.88 (0.50) & 88.83 (0.35) & 96.48 (0.50) & \textcolor{blue}{9.60} & \textcolor{blue}{85.63}  
& 98.71 (0.14) & 100.00 (0.00) & 94.31 (0.13) & 100.00 (0.00) & \textcolor{blue}{2.00} & \textcolor{blue}{94.47}  \\

RL 
& 92.32 (0.67) & 96.86 (0.64) & 86.81 (0.39) & 97.25 (0.54) & \textcolor{blue}{11.76} & \textcolor{blue}{82.41}  
& 98.08 (0.16) & 100.00 (0.00) & 93.93 (0.11) & 100.00 (0.00) & \textcolor{blue}{2.71} & \textcolor{blue}{93.88}  \\

IU 
& 93.19 (5.05) & 75.06 (42.82) & 87.06 (5.20) & 78.61 (36.98) & \textcolor{blue}{34.42} & \textcolor{blue}{61.21}  
& 96.26 (2.45) & 75.00 (43.30) & 91.49 (2.92) & 96.62 (5.86) & \textcolor{blue}{25.87} & \textcolor{blue}{71.14}  \\

BE 
& 98.62 (0.29) & 83.63 (0.93) & 92.33 (0.06) & 99.39 (0.08) & \textcolor{blue}{16.61} & \textcolor{blue}{77.19}  
& 99.51 (0.18) & 97.09 (0.24) & 95.66 (0.02) & 100.00 (0.00) & \textcolor{blue}{2.96} & \textcolor{blue}{94.86}  \\

BS 
& 98.60 (0.55) & 76.93 (16.82) & 92.47 (0.69) & 98.24 (2.01) & \textcolor{blue}{23.28} & \textcolor{blue}{69.87}  
& 99.52 (0.18) & 79.89 (23.95) & 95.49 (0.05) & 100.00 (0.00) & \textcolor{blue}{20.12} & \textcolor{blue}{75.93}  \\

$\ell_{1}$-sparse 
& 88.58 (1.82) & 100.00 (0.00) & 84.93 (1.79) & 100.00 (0.00) & \textcolor{blue}{15.02} & \textcolor{blue}{79.73}  
& 93.26 (0.61) & 100.00 (0.00) & 92.81 (0.58) & 100.00 (0.00) & \textcolor{blue}{7.39} & \textcolor{blue}{88.26}  \\

SalUn 
& 95.08 (0.64) & 99.22 (0.37) & 89.13 (0.28) & 99.31 (0.33) & \textcolor{blue}{7.49} & \textcolor{blue}{88.25}  
& 98.22 (0.15) & 100.00 (0.00) & 93.94 (0.13) & 100.00 (0.00) & \textcolor{blue}{2.61} & \textcolor{blue}{94.02}  \\

CUP 
& 97.00 (0.58) & 97.36 (1.05) & 90.82 (0.43) & 97.89 (0.88) & \textcolor{blue}{\textbf{5.95}} & \textcolor{blue}{\textbf{90.41}}  
& 99.50 (0.20) & 99.89 (0.12) & 95.07 (0.12) & 99.98 (0.03) & \textcolor{blue}{\textbf{0.93}} & \textcolor{blue}{\textbf{95.18}}  \\

\bottomrule
\end{tabular}
}
\end{table*}

\begin{table*}[h]
\setlength{\tabcolsep}{2pt}
\scriptsize \centering
\caption{Class-wise unlearning performance in image classification task on CIFAR-10 and SVHN (class 9). The reported values represent the average performance across 5 different random seeds. The value within parentheses indicates the standard deviation.}
\resizebox{\textwidth}{!}{
\begin{tabular}{lcccccccccccc}
\toprule
\multirow{2}{*}{\textbf{Methods}} & \multicolumn{6}{c}{\textbf{CIFAR-10}} & \multicolumn{6}{c}{\textbf{SVHN}} \\ 
\cmidrule(lr){2-7} \cmidrule(lr){8-13}
 & RA ($\uparrow$) & UA ($\uparrow$) & TA ($\uparrow$) & MIA ($\uparrow$) & \(\Delta (\downarrow)\) & \(\mathcal{H} (\uparrow)\)  
 & RA ($\uparrow$) & UA ($\uparrow$) & TA ($\uparrow$) & MIA ($\uparrow$) & \(\Delta (\downarrow)\) & \(\mathcal{H} (\uparrow)\)  \\ 
\midrule
Retrain 
& 100.00 (0.00) & 100.00 (0.00) & 94.88 (0.51) & 100.00 (0.00) & \textcolor{blue}{0.00} & \textcolor{blue}{94.88}  
& 100.00 (0.00) & 100.00 (0.00) & 94.85 (0.63) & 100.00 (0.00) & \textcolor{blue}{0.00} & \textcolor{blue}{94.85} \\ 
\midrule

GA 
& 97.43 (0.29) & 94.49 (0.37) & 90.72 (0.11) & 97.63 (0.19) & \textcolor{blue}{7.66} & \textcolor{blue}{85.35}  
& 99.50 (0.20) & 98.48 (0.09) & 89.44 (0.05) & 99.79 (0.04) & \textcolor{blue}{1.76} & \textcolor{blue}{88.88}  \\

WS 
& 95.84 (0.40) & 97.10 (0.36) & 89.61 (0.28) & 97.70 (0.28) & \textcolor{blue}{7.56} & \textcolor{blue}{88.07}  
& 99.09 (0.14) & 99.78 (0.04) & 88.78 (0.15) & 99.91 (0.04) & \textcolor{blue}{1.65} & \textcolor{blue}{90.84}  \\

RL 
& 93.18 (0.50) & 98.32 (0.18) & 87.36 (0.21) & 98.62 (0.17) & \textcolor{blue}{10.27} & \textcolor{blue}{84.35}  
& 98.51 (0.16) & 99.90 (0.06) & 88.29 (0.22) & 99.95 (0.05) & \textcolor{blue}{2.38} & \textcolor{blue}{89.83}  \\

IU 
& 91.25 (5.85) & 75.13 (43.05) & 85.64 (5.77) & 78.02 (38.08) & \textcolor{blue}{35.51} & \textcolor{blue}{55.93}  
& 99.17 (0.50) & 75.00 (43.30) & 89.94 (3.19) & 98.52 (2.57) & \textcolor{blue}{25.06} & \textcolor{blue}{79.24}  \\

BE 
& 98.85 (0.29) & 91.85 (0.58) & 92.25 (0.08) & 99.85 (0.04) & \textcolor{blue}{8.59} & \textcolor{blue}{84.72}  
& 99.64 (0.18) & 97.39 (0.27) & 90.09 (0.02) & 100.00 (0.00) & \textcolor{blue}{2.63} & \textcolor{blue}{89.51}  \\

BS 
& 99.11 (0.35) & 81.70 (21.40) & 92.86 (0.56) & 99.72 (0.23) & \textcolor{blue}{18.42} & \textcolor{blue}{75.68}  
& 99.63 (0.19) & 80.97 (23.44) & 90.83 (1.03) & 100.00 (0.00) & \textcolor{blue}{19.05} & \textcolor{blue}{74.23}  \\

$\ell_{1}$-sparse 
& 88.87 (1.24) & 100.00 (0.00) & 85.06 (1.08) & 100.00 (0.00) & \textcolor{blue}{14.74} & \textcolor{blue}{79.83}  
& 99.95 (0.02) & 100.00 (0.00) & 89.70 (0.02) & 100.00 (0.00) & \textcolor{blue}{0.44} & \textcolor{blue}{89.66}  \\

SalUn 
& 97.92 (0.30) & 96.77 (0.44) & 91.70 (0.21) & 97.64 (0.42) & \textcolor{blue}{5.43} & \textcolor{blue}{90.17}  
& 98.67 (0.20) & 99.81 (0.04) & 88.49 (0.19) & 99.95 (0.02) & \textcolor{blue}{2.13} & \textcolor{blue}{90.33}  \\

CUP 
& 98.34 (0.23) & 97.94 (0.98) & 91.67 (0.18) & 98.48 (0.75) & \textcolor{blue}{\textbf{4.32}} & \textcolor{blue}{\textbf{92.04}}  
& 99.70 (0.17) & 99.63 (0.25) & 89.68 (0.09) & 99.94 (0.09) & \textcolor{blue}{\textbf{0.67}} & \textcolor{blue}{\textbf{91.78}}  \\

\bottomrule
\end{tabular}
}
\end{table*}

\begin{table}[!htb]
\setlength{\tabcolsep}{2pt}
\caption{Unlearning performance for the image classification task on CIFAR-10 (\(10(\%)\) random subset forgetting). Reported values are computed over 5 different random seeds, with standard deviations shown in parentheses.}
\centering
\small
\begin{tabular}{lccccc}
\toprule
\textbf{Methods} & RA ($\uparrow$) & UA ($\uparrow$) & TA ($\uparrow$) & MIA ($\uparrow$) & \(\Delta (\downarrow)\) \\
\midrule
Retrain & 100.00 (0.00) & 5.32 (0.56) & 94.27 (0.14) & 13.26 (0.72) & \textcolor{blue}{0.00}  \\
\midrule
GA    & 99.40 (0.44) &  0.81 (0.46) & 94.14 (0.65) & 1.59 (0.84) & \textcolor{blue}{12.53}  \\
WS    & 99.53 (0.25) &  1.10 (0.56) & 93.84 (0.37) & 3.07 (1.07) & \textcolor{blue}{11.05}  \\
RL    & 99.58 (0.21) &  0.71 (0.36) & 94.48 (0.17) & 1.37 (0.48) & \textcolor{blue}{12.76}  \\
IU    & 95.18 (4.04) &  4.76 (3.84) & 89.39 (3.87) & 7.60 (5.20) & \textcolor{blue}{8.91}  \\
BS    & 94.35 (1.73) &  5.61 (2.00) & 88.08 (1.89) & 13.43 (1.27) & \textcolor{blue}{8.38}   \\
BE    & 95.93 (0.73) &  3.93 (0.78) & 89.61 (0.68) & 23.50 (0.46) & \textcolor{blue}{12.05}  \\
WS    & 99.53 (0.25) &  1.10 (0.56) & 93.84 (0.37) & 3.07 (1.07) & \textcolor{blue}{11.05}   \\
$\ell_{1}$-sparse     & 99.76 (0.11) &  0.36 (0.18) & 94.59 (0.07) & 2.99 (0.31) & \textcolor{blue}{11.40}   \\
SalUn & 98.87 (0.68) &  1.94 (1.01) & 93.27 (0.87) & 3.14 (1.32) & \textcolor{blue}{10.78}  \\
\midrule
CUP   & 97.43 (1.49) & 5.78 (3.22) & 91.02 (1.92) & 8.79 (4.34) & \textcolor{blue}{\textbf{6.11}}  \\
\bottomrule
\end{tabular}\label{tab:random}
\end{table}

\end{document}